\RequirePackage{amsthm}
\let\qed\relax
\documentclass[preprint,12pt]{elsarticle}
\usepackage{graphicx}%
\usepackage{multirow}%
\usepackage{amsmath,amssymb,amsfonts}%
\usepackage{mathrsfs}%
\usepackage[title]{appendix}%
\usepackage{xcolor}%
\usepackage{textcomp}%
\usepackage{manyfoot}%
\usepackage{booktabs}%
\usepackage{lmodern}
\usepackage{listings}%
\usepackage{hyperref} 
\usepackage{subcaption}
\usepackage{svg}
\usepackage{gensymb}
\usepackage[ruled,vlined]{algorithm2e}
\usepackage[british]{babel}
\usepackage{csquotes}
\usepackage{amsthm}%
\usepackage{amsmath,amsfonts,bm}

\usepackage{letltxmacro}
\LetLtxMacro{\originaleqref}{\eqref}

\def\eqref#1{equation~\originaleqref{#1}}
\def\1{\bm{1}}

\def\vtheta{{\bm{\theta}}}

\def\vf{{\bm{f}}}

\def\vk{{\bm{k}}}

\def\vx{{\bm{x}}}
\def\vy{{\bm{y}}}

\def\mI{{\bm{I}}}

\def\mK{{\bm{K}}}

\DeclareMathAlphabet{\mathsfit}{\encodingdefault}{\sfdefault}{m}{sl}
\SetMathAlphabet{\mathsfit}{bold}{\encodingdefault}{\sfdefault}{bx}{n}

\def\sH{{\mathbb{H}}}

\journal{Materials \& Design}

\begin{document}

\begin{frontmatter}



\title{Machine learning for in-situ composition mapping in a self-driving magnetron sputtering system}

\author[label1,label2]{Sanna Jarl\corref{mycorrespondingauthor}}
\cortext[mycorrespondingauthor]{Corresponding author}
\ead{sanna.jarl@angstrom.uu.se}

\author[label3]{Jens Sjölund}
\ead{jens.sjolund@it.uu.se}

\author[label4]{Robert J. W. Frost}
\ead{rob.frost@physics.uu.se}

\author[label2]{Anders Holst}
\ead{anders.holst@ri.se}

\author[label1]{Jonathan J. S. Scragg}
\ead{jonathan.scragg@angstrom.uu.se}

\affiliation[label1]{organization={Materials Science and Engineering, Uppsala University},
            addressline={Regementsvägen 1},
            city={Uppsala},
            postcode={75237},
            state={},
            country={Sweden}}

\affiliation[label2]{organization={Computer Science, RISE Research Institutes of Sweden},
            addressline={Isafjordsgatan 22},
            city={Kista},
            postcode={16440},
            state={},
            country={Sweden}}
            
\affiliation[label3]{organization={Information Technology, Uppsala University},
            addressline={Regementsvägen 1},
            city={Uppsala},
            postcode={75237},
            state={},
            country={Sweden}}
            
\affiliation[label4]{organization={Physics and Astronomy, Uppsala University},
            addressline={Regementsvägen 1},
            city={Uppsala},
            postcode={75237},
            state={},
            country={Sweden}}

%% Abstract
\begin{abstract}
Self-driving labs (SDLs) employing automation and machine learning (ML) offer great promise for accelerating materials discovery and optimisation. However, in thin film science, SDLs are mainly restricted to solution-based methods which are easier to automate, restricting access to the broader chemical space of inorganic materials. This work advances an SDL based on magnetron co-sputtering, addressing a key challenge: rapidly generating accurate composition maps of multi-element, compositionally graded thin films. Traditional ex-situ methods are slow and error-prone; instead, we present a fast, calibration-free, in-situ ML approach to predict the deposition rate using quartz-crystal microbalance (QCM) sensors. For each sputtering source, deposition rates are sequentially learned as a function of pressure and power via active learning with Gaussian processes (GPs). The final GPs are combined with a geometric flux model to interpolate deposition rates across the sample. Among several acquisition functions with random query as the baseline, the Bayesian active learning MacKay (BALM) approach yielded the best performance, requiring as few as 10 experiments per source. The model predictions for co-sputtering composition distributions were validated against external composition measurements. This framework significantly increases throughput in combinatorial sputtering studies and highlights the potential of ML-guided SDLs to surpass traditional Edisonian methods.
\end{abstract}

%% Keywords

\end{frontmatter}


\section{Introduction}\label{sec:intro}

% Generic intro
Identifying and developing novel materials with specific properties, structures, and functionalities is essential to advance technology, improve existing products, and address future challenges. 
% Conventional approach
The development of new materials requires understanding (or at least, determination) of their synthesis-property relationships (SPRs). For materials of any complexity, this is rarely obtainable from first principles, and instead must be derived experimentally. This entails performing many sequential cycles of material synthesis and characterisation, in an open-ended attempt to first establish empirical SPRs, and ultimately to discover their underlying physical origins \citep{yao2023machine}. Given the possible combinations of elements, the search space for new materials is vast \citep{merchant2023scaling}. At the same time, any chosen method of synthesis introduces an additional set of variable parameters, creating an exponentially increasing search space for experimentalists.   

% Data driven approach
Now, advances in machine learning (ML) are leading a transition in materials science toward a data-driven approach, where the exploration of novel materials and prediction of material properties have been among the most active fields of research \citep{chibani2020machine, gubernatis2018machine}. Moreover, the integration of ML, lab automation, and robotics has resulted in the advent of self-driving labs (SDLs) \citep{abolhasani2023rise}, greatly accelerating the discovery process in applications such as pharmaceuticals \citep{schneider2018automating, roberts2025automating} and batteries \citep{yik2023automated}. A key component in an SDL is the ML-driven design of experiments. Whether the SDL is tasked with optimizing the performance of a material with respect to a well-defined metric, or in maximizing the information gain of each experiment, Bayesian optimization \citep{garnett2023bayesian} offers a convenient framework \citep{alghalayini2025machine,hickman2022bayesian, yik2025accelerating,liang2021benchmarking}. This enables SDLs to specialize towards either materials optimisation in more applied or commercial contexts, or materials exploration, i.e. characterisation of SPRs, when the underlying materials science is of primary interest.

Thin film materials are a useful medium for high-throughput materials discovery in SDL contexts or otherwise. So far, most examples of thin film SDLs use solution coating methods for film fabrication, and some highly advanced systems exist. Key examples include systems aimed at the development of functional organic and perovskite solar cells and their constituent layers \citep{zhang2024toward, macleod2021advancing}. In contrast, SDLs applying vacuum-based methods, Physical Vapour Deposition (PVD) or Chemical Vapour Deposition, are far less developed. This is presumably due to additional challenges of safe, high-throughput hardware automation, lack of standardised components, and greater costs involved compared to solution processing. Published examples of PVD-based SDLs include a system for optimising the reflectivity and absorptivity of sputtered single-element films \citep{zheng2024machine} and another for minimising the resistance of reactively sputtered Nb-doped TiO$_2$ deposited from individual compound targets\citep{shimizu2020autonomous}. \cite{harris2024autonomous} used pulsed laser deposition to grow of WSe$_2$ using in-situ Raman spectroscopy to estimate the growth parameters needed to maximize the film crystallinity. Further examples include \cite{wakabayashi2019machine, wakabayashi2022bayesian} and \cite{ohkubo2021realization}, where SrRuO$_3$ respectively TiN thin films were prepared using molecular beam epitaxy to obtain desired resistivity or magnetic anisotropy, or optimizing crystallinity, based on in-situ reflection high-energy electron diffraction. We note that all these examples used single material sources, with one stage of synthesis. An example using two sources was provided by \cite{febba2023autonomous}, where Bayesian optimisation was used to predict composition of Zn$_x$-Ti$_{1-x}$-N films deposited by reactive sputtering from Zn and Ti targets. In order to tune the power for each target, they performed in-situ plasma characterisation by optical emission spectroscopy. This method is widely applicable, although it does require external calibration.

For broader materials exploration, PVD-based SDLs need to combine multiple material sources to generate arbitrary material compositions. They should also take advantage of the powerful workflows of high-throughput thin film science, in which a common technique is to grow samples with graded compositions (combinatorial samples), to obtain information on the effect of composition on material properties in parallel with varying other synthesis parameters \citep{ludwig2019discovery}. Here, accurate knowledge of the composition distribution over the thin film is crucial, but is difficult to integrate into an SDL. One possibility is to include in-line characterisation tools such as Energy-dispersive X-ray Spectroscopy (EDS) or X-ray fluorescence (XRF). However besides being very time-consuming, and thus limiting sample throughput, these methods suffer from systematic errors and matrix effects \citep{rousseau2006corrections}, thus requiring calibration standards which will generally not be available for novel materials. Absolute, standard-free composition mapping can be performed via e.g., Rutherford Backscattering Spectrometry (RBS)\citep{RBS}, but this method is not feasible to integrate in an SDL as it requires an extensive experimental infrastructure of its own.

In this work, we present a developing SDL based on magnetron sputtering, which combines in-situ sensing, machine learning and process modelling to provide accurate composition maps for combinatorial samples in real time, without the need for external calibration. We validate the method against RBS measurements, and show how, in an SDL context, the approach can be used to automatically define process conditions needed to obtain specified compositions, for arbitrary combinations of material sources. These features will be critical to performing high-throughput automated materials exploration in a thin film PVD-based SDL.

% Now to the specifics of our case
The basis for inference of composition gradients across the combinatorial thin films is characterisation of the sputter flux distribution arising from each magnetron sputtering source. The sources emit material in a directed \enquote{plume} that spreads into the chamber, with an intensity and distribution that depends on the power applied to the magnetron and the chamber plasma pressure. Therefore, these variables will influence the deposition rate over the sample and the resulting composition and composition gradients on a static sample during deposition. In principle, the flux distribution might be predicted based on physical models of the sputtering process, such as developed by \cite{complex-flux-model}. However, speaking to the complexity of the processes involved, such models do not provide exceptionally good fits to experimental data. Simpler, geometrical models have been described that better match experimental results, based on several adjustable parameters, though without any explicit connection to physical variables such as chamber pressure or applied power \citep{original-flux-model}. 

\begin{figure}[t]
    \centering
    \includegraphics[width=\linewidth]{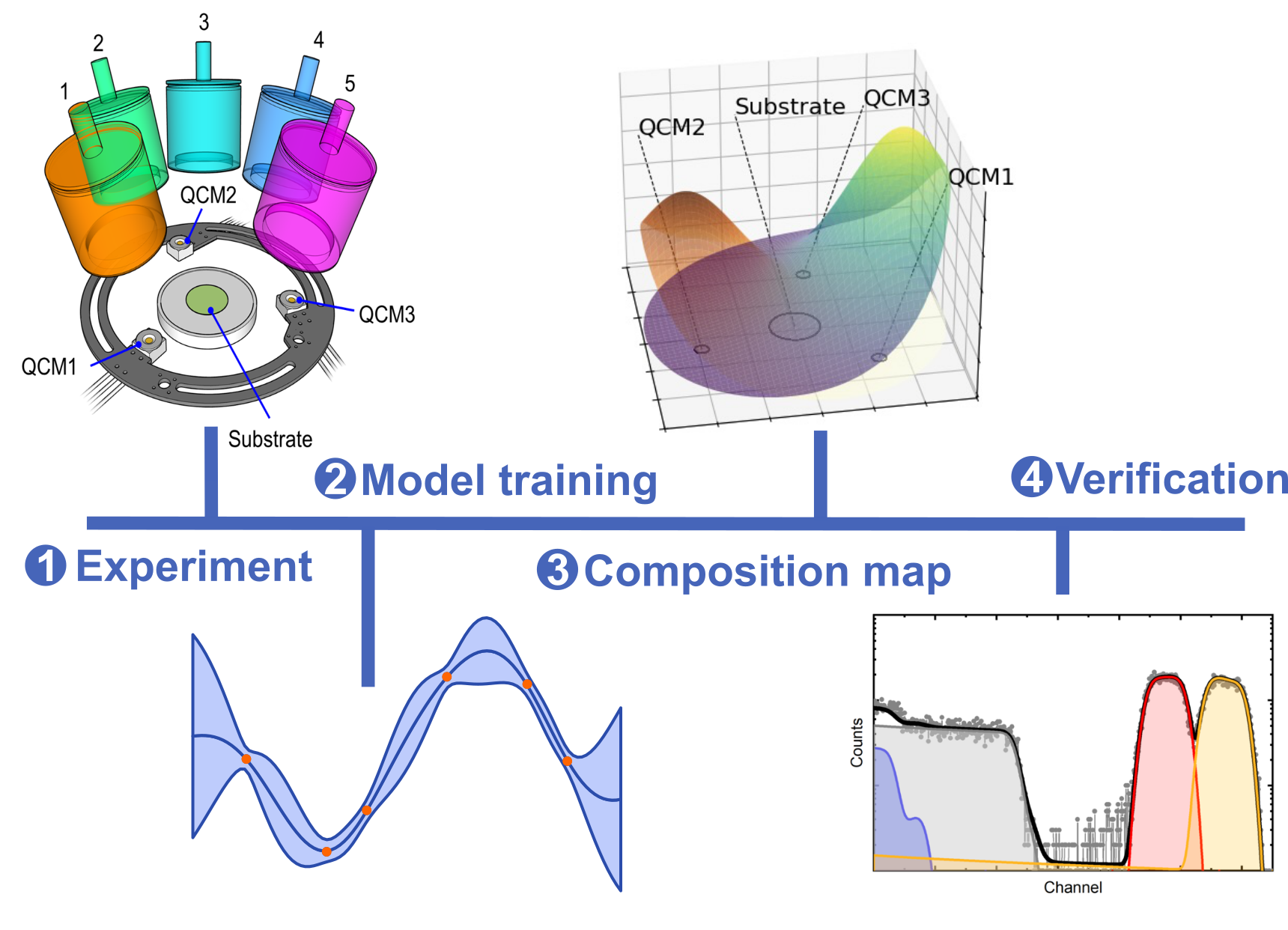}
    \caption{The SDL workflow consists of the following steps each iteration: 1. PVD sputtering experiment, 2. model training with active learning, 3. obtain composition map to infer co-sputtering. Finally, external verification by RBS.}
    \label{fig:workflow}
\end{figure}

For measuring sputter flux in-situ, it is common to use quartz-crystal microbalance (QCM) sensors, which monitor changes in resonant frequency of a thin quartz crystal as material is deposited upon it, to determine the mass accumulation upon them. Multiple, and even movable sensors, can also be used to perform 3D profiling of the sputter flux plume by sweeping a set of three QCMs across the deposition zone \citep{suram2015combinatorial}. In our SDL we situate three QCMs at fixed locations around the sample, and use a geometrical flux model to interpolate the deposition rate to at other locations -- specifically the ${(x,y)}$ coordinates across the area of the substrate. While QCMs are highly accurate, they only measure the total mass-deposition rate, without distinguishing contributions from different sources. Thus it is necessary to characterise the fluxes from individual sources first, before combining the results to predict the outcome of multi-source sputtering. Flux characterisation from single sources is achieved by running a sequential learning process in the SDL, training a machine learning model, in this study a Gaussian process (GP), to predict the QCM readings across the entire power-pressure parameter space for the given source. Different acquisition functions were investigated, to minimise the number of experiments needed to train for each source. This is critical to minimise source material wastage and to be able to train and retrain models as the targets are consumed over time. Trained models for several sources are then combined to predict sample composition maps, which are validated against ex-situ characterisation of samples produced by the SDL, using RBS.  See Figure \ref{fig:workflow} for an overview of the workflow in the SDL. 

\section{Methods}\label{sec:methods}
In this section we present the SDL experimental setup and automation, process modelling and the procedure for training the machine learning models and experimental validation. 

\subsection{Self-Driving Lab (SDL) equipment and operation} 
Our SDL is based on a Kurt J. Lesker sputter chamber with six magnetrons oriented toward the substrate location, at a target-substrate distance of 155 mm. Three water cooled QCMs with polished Au electrodes have been installed around the substrate holder at 120 degree intervals, as shown in the top left part of Figure \ref{fig:workflow}.  We automated the sputtering system using a custom Python script utilizing pywinauto\footnote{https://github.com/pywinauto/pywinauto} that interacts with the system via the manufacturer's Eklipse graphical user interface. This approach enables automation in the absence of an API, meanwhile avoiding unsafe operation of the system. The script is capable of sequencing operations of the system from startup to shutdown, performing series of experiments (with or without sample production), either from predetermined recipes or under algorithmic control. It records all metadata and measured data and stores it in a MongoDB database. The automated workflow for characterising sputter flux distribution is based only on QCM feedback, without the need to deposit any films. It begins with preparing the chamber by adjusting gas injection and pumping rates to the desired initial setpoints, and pre-sputtering the targets. Then, a sequence of experiments is initiated, in which the system varies the powers $P_k$, supplied to each of several magnetrons $k$, and the argon pressure $p_\text{Ar}$ in the sputtering chamber, where these conditions are being chosen by an acquisition function (see below). For each experimental point, the applied conditions are first stabilised, and then the QCM shutters are held open for 20 s to obtain mass deposition rate readings. 

\subsection{Sputter flux model} \label{sec:flux_model}
The rate of change of measured frequency $\dot{f}_{c,i}$ at QCM $i\in\{1,2,3\}$  is converted to a mass deposition rate $\dot{m}_i$ by taking the time derivative of the Sauerbrey equation \citep{sauerbrey}, yielding
\begin{equation}
    \dot{m}_i = - \frac{\sqrt{\mu\rho}}{2f^2}\cdot \dot{f}_{c,i}
\end{equation}
where $\rho$ is the quartz density, $\mu$ is the shear modulus for AT-cut quartz, $f$ is the frequency of the unloaded quartz crystal, and $f_{c}$ is the measured frequency. To stay well within the applicability range of the Sauerbrey equation - where the deposited mass should be small compared to the mass of the crystal - we replaced the quartz crystals once their frequency dropped below 97.5\% of the original value, 6MHz. The QCMs are water cooled to maintain constant temperature, with polished gold coating. The error in $\dot{m}_i$ resulting from the measurement and application of the Sauerbrey equation is approximately 5\%.

\begin{figure}
    \centering
    \includegraphics[width=0.5\linewidth]{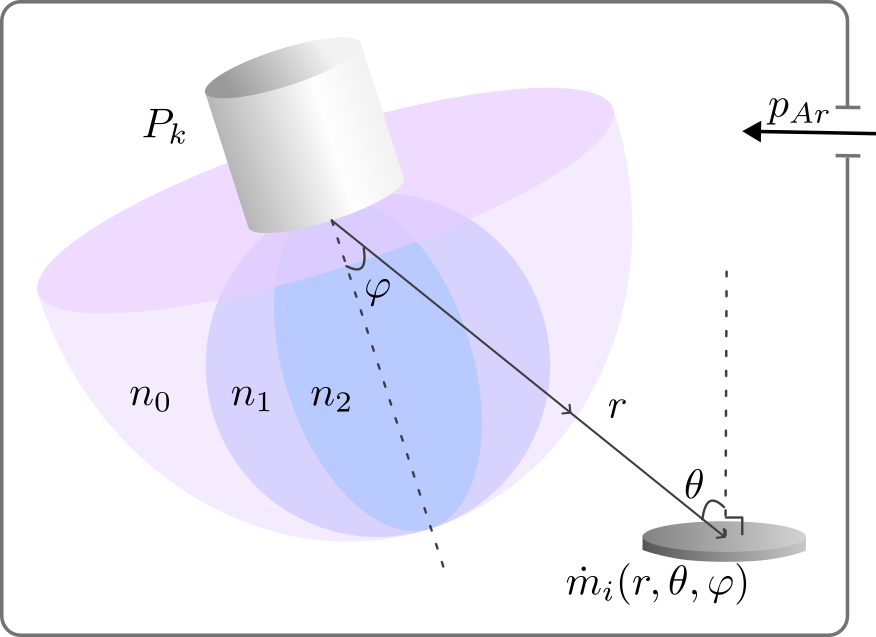}
    \caption{Geometric model of the sputter flux. The sputter gun, which emits matter, is shown in grey. Isosurfaces with equal concentrations of emitted atoms are shown in different shades. The emission becomes increasingly directed and narrow as $n$ increases.}
    \label{fig:flux_model}
\end{figure}

To infer the sputter flux $\dot{m}_{x,y,z}$ at other coordinates $(x,y,z)$ between the QCM sensors, a geometric flux model according to \cite{wasa2012handbook} is used. In which the distribution of sputtered atoms falls off as approximately $\cos^n \varphi$ where $\varphi$ is the angle from the sputter-source surface normal shown in Figure \ref{fig:flux_model}. 
Based on this, the rate of deposition follows
\begin{equation}
    \ \dot{m} = \frac{a (n+1) \cos \theta \cos^n \varphi}{2\pi r^2},
    \label{eq:flux}
\end{equation}
where $a$ is the emission rate from the source, $n$ is the order of the cosine distribution describing the shape of the flux, $r$ is the distance from the centre of the magnetron target surface to the point of deposition, and $\theta$ is the incident angle between the normal of the receiving surface (QCM or sample) and the vector from the magnetron centre, see Figure \ref{fig:flux_model}. Values for $\theta$, $\varphi$  and $r$ are computed based on careful measurements of the chamber's internal geometry. At the substrate centre, typical values are $\theta$ = 43.2\textdegree, $\varphi$ = 28.4\textdegree and $r$ = 155 mm.
For a given sputter process determined by setpoints of $P_k$ and $p_\text{Ar}$, $n$ and $a$ values are fitted by minimising the root mean square difference between the experimental readings $\dot{m}_i$ at the three sensors and the flux model predictions at the corresponding QCM positions. This procedure is detailed in \cite{david-sorme-thesis}. These calculations are implemented in the SDL code and performed automatically during operation, resulting in characterisation of the specific flux distribution for each sputtering condition tested. 

\subsection{Active learning framework}\label{sec:active_learning}
\begin{figure}
    \centering
    \includegraphics[width=\linewidth]{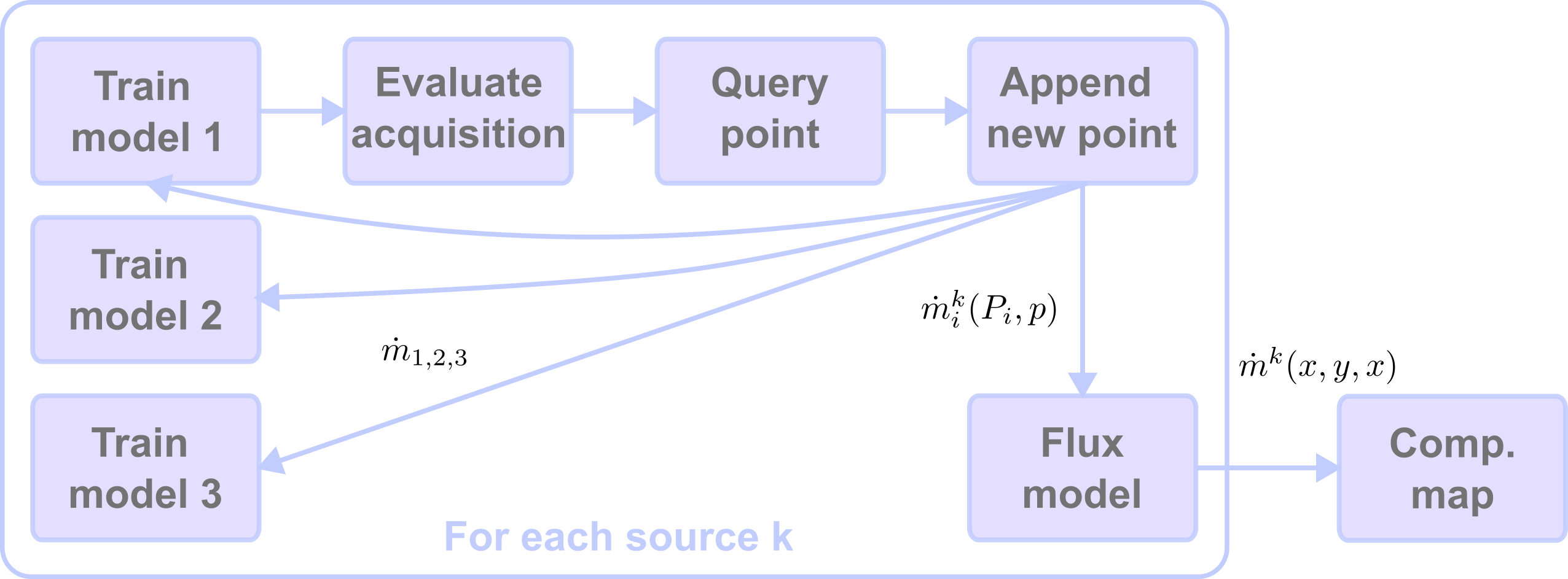}
    \caption{The active learning cycle begins with training three models (one for each QCM sensor mass rate) on several initial experiments, then iteratively evaluating the acquisition function and querying the next experimental point, appending the new point to the training set and repeating until the budget is spent. After training, the three sensor models are used to fit a flux distribution model (described in Section \ref{sec:flux_model}) for the source. Subsequently, multiple such flux models can be combined to predict composition maps for co-sputtering depositions.}
    \label{fig:AL_flow}
\end{figure}

We perform active learning via Bayesian Optimisation (BO), where a surrogate model (GP) is maintained and used to select the next experiment by minimising the value of an acquisition function, given the current model. The outcome of the experiment is added to the dataset and used to update the surrogate model, after which the process is repeated. The full active learning procedure is outlined in Figure \ref{fig:AL_flow}. 
For the purposes of developing the active learning framework, we used a reference dataset where the mass deposition rates $\dot{m}_i$ were measured on a uniform grid of power $P_k$ and pressure $p_{Ar}$, for the specific case of a Zr sputtering target (the methodology is transferable to any target material). The ranges of $P_{Zr}$ and $p_{Ar}$ were 1--43 W in 3 W steps and 1--43 mTorr in 3 mTorr steps for a total of 225 datapoints, which required approximately 2 hours of sputtering time in the SDL. 

In each active learning experiment, the model was initialised by an initial set of five points, either randomly or deliberately chosen, with another 50 being sequentially acquired during active learning. The BO parts of the workflow were implemented in Botorch \citep{balandat2020botorch}, which uses GPyTorch \citep{gardner2018gpytorch} as the computational backend for the GP computations. Our code and training dataset is available on GitHub \footnote{https://github.com/jarlsanna/gps-for-magnetron-sputtering\#}.

If the argon pressure inside the chamber becomes too low, the plasma extinguishes and sputtering cannot occur, which gives rise to a sudden drop in the deposition rate at low pressures. Since we are not interested in such regions and it is a common phenomenon for any metal source, this area can safely be filtered out by setting a lower limit on the pressure $p_{Ar}$, in this case a constant limit of 4 mTorr is used. However, ideally one would find the lower bound for each source as it depends on the material and the magnetron power. 

To efficiently learn the parameter space in a minimum number of experiments, three different ways of restricting the ($P_{Zr}$, $p_{Ar}$) search space and choosing the initial training points were studied. \textbf{Case 1: }The naive case, with quasi-random (Sobol) sampled initial points, from the complete search space. \textbf{Case 2: }Sobol sampling of initial points, while restricting the search space by setting  $p_{Ar}$ to a minimum of 4 mTorr. This avoids the sudden drop-off in deposition rate due to plasma quenching. \textbf{Case 3: }Pre-defined initial points, taking the 4 corner points and the centre point in ($P_{Zr}$, $p_{Ar}$), and restricting the search space like in Case 2. This is motivated by the desire to span the entire input space, and is reliable because similar behaviour for any target is expected, i.e., the same physical processes are occurring with quantitative rather than qualitative differences. Thus, predefining the initial points is a potentially pragmatic way to reduce the training budget.

For each target, single-source experiments were performed to train individual models, following the procedure in Figure \ref{fig:AL_flow}. In real time, we only learn the deposition rate $\dot{m}$ in \eqref{fig:flux_model} from the QCM sensor located closest to the target, since this gives the strongest and most stable signal. From a physical perspective the deposition rate measured by all sensors should exhibit similar behaviour due to a common flux source. Therefore, the data points acquired for one sensor can also inform models for the others enabling rate predictions. Note that the other two models are not used during active learning. Since the sputter flux model described in Section \ref{sec:flux_model} needs the deposition rate from all sensors, the additional models are used to obtain deposition rate predictions, which could be done offline after all experiments are performed.

To evaluate the model performance, we calculate the root mean square error ($E_\text{RMS}$) at each iteration. $E_\text{RMS}$ is given by
\begin{equation}
    E_\text{RMS} = \sqrt{\frac{\sum_{i=1}^N ||y_i-\hat{y}_i||^2}{N}}
    \label{eq:rmse}
\end{equation}
where $y$ and $\hat{y}_i$ are the observations respectively predictions, using our grid dataset. To quantify how well the active learning process performs, we use the enhancement factor in terms of $E_\text{RMS}$ \citep{rohr2020benchmarking}, defined by
\begin{equation}
    \text{F} = \frac{E_\text{RMS}^\text{baseline}(j)}{E_\text{RMS}^{\text{method}}(j)}
    \label{eq:ef}
\end{equation}
where $E_\text{RMS}^\text{baseline}(j)$ is the $E_\text{RMS}$ at iteration $j$ for the baseline which in our case is random queries, and $E_\text{RMS}^\text{method}(j)$ is the corresponding $E_\text{RMS}$ score for the method evaluated. If the enhancement factor is greater than 1, it means the active learning process results in better model fits compared to the (random) baseline. Since some of the active learning methods evaluated are stochastic, we repeated each experiment 10 times to account for randomness and different initialisations.

Once the framework was evaluated, and a suitable model and acquisition function selected, we deployed it in the SDL to learn sputtering deposition rates in real time. A fixed number of iterations (30) was used to train a GP for each magnetron source of interest.

\subsection{Gaussian processes}
A common choice of surrogate model is a Gaussian process, which defines a distribution over functions that is conditioned to agree with observation data $\mathcal{D} = \{\vx_i, y_i \}_{i=0}^N$. Typically, the unknown function of interest $\vf$ is assumed to be perturbed with independent and identically distributed Gaussian noise $\varepsilon \sim N(0, \sigma_n^2)$, i.e. the measurement at $\vx_i$ is given by $y_i = f_i + \varepsilon_i$ where $f_i=f(\vx_i)$. A GP is characterised by a prior mean $\mu(\vx)$ and a positive definite covariance kernel $k(\vx,\vx')$, where the covariance kernel captures structural assumptions such as smoothness and periodicity \citep{williams2006gaussian}. 

Given observation data, the posterior mean and variance at a new location $\vx_*$ can be expressed in closed form as 
\begin{align}
    \mu_n(\vx_*) &=  \mu(\vx_*) + \vk_n(\vx_*)^{\top}(\mK_n + \sigma_n^2 \mI)\vy, \\
    s_n^2(\vx_*) &= k(\vx_*, \vx_*) - \vk_n(\vx_*)^{\top}(\mK_n + \sigma_n^2 \mI)\vk_n(\vx_*),
\end{align}
where $(\mK_n)_{ij} = k(\vx_i, \vx_j), \vk_n(\vx) = [k(\vx, \vx_1),...,k(\vx, \vx_n)]^{\top}$ and $\sigma_n^2$ is the observation noise variance of $f$. Point estimates of the hyperparameters are determined by maximising the marginal log-likelihood, which is available in closed form.

Gaussian processes can, however, be extended to account for hyperparameter uncertainty via a fully Bayesian approach, i.e. placing a prior $p(\vtheta)$ over the hyperparameters and marginalizing the full posterior over the hyperparameters. In order to make predictions at new locations $\vx_*$, we compute the predictive posterior over $\vf_*$ given the training data $\vy$ and integrate out both the latent function values $\vf_*$ and the hyperparameters $\vtheta$. The integral is intractable and needs to be approximated with, for instance, a Monte Carlo (MC) method. Omitting the conditioning over inputs for brevity, the resulting MC estimate of the predictive posterior is 
\begin{equation}
    p(\vf_* \mid \vy) \approx \frac{1}{M}\sum_{j=1}^Mp(\vf_* \mid \vy, \vtheta_j), \quad \vtheta_j \sim p(\vtheta \mid \vy), \quad j=1,\ldots, M.
\end{equation}

In this work, we compare a standard GP with a fully-Bayesian one. In both cases we use a Matérn 5/2 kernel with constant mean. An rbf kernel was also evaluated which gave similar results, but the Matérn kernel was chosen to stay more flexible. In the fully Bayesian case, after studying the convergence, we used 32 warm-up steps and took 1024 MC samples without any thinning. To model feature importance, Automatic Relevance Determination is used, meaning each input dimension is individually scaled by lengthscales $\ell_i$ \citep{williams2006gaussian}. In $D$ dimensions the distance between points $\vx$ and $\vx'$ is computed as $r^2 = \sum_{i=1}^D(x_i-x'_i)^2/\ell_i^2$. This gives us the trainable parameters $\vtheta = \{ \mathbf{\ell}, \sigma_n, \sigma_f, c \}$, with $\sigma_f$ being the signal variance and $c$ the constant mean.
The fully Bayesian model uses a sparse axis-aligned subspaces prior \citep{eriksson2021high}, inducing a sparse structure in the inverse squared length scales $\rho_i \sim \mathcal{HC}(\tau)$, where $\mathcal{HC}$ denotes the half-Cauchy distribution and $\tau \sim \mathcal{HC}(\alpha)$ with $\alpha > 0$ being a scalar parameter controlling the level of shrinkage. The kernel variance $\sigma_k^2 \sim \mathcal{LN}(0, 10^2)$ is a log-Normal distribution.

\subsection{Acquisition function}
The purpose of the acquisition function in active learning is to choose the most informative point to evaluate and thereby minimise the number of experiments necessary \citep{settles2009active}. It is crucial to select a suitable acquisition function reflecting the problem character, in our case to learn the entire parameter space rather than optimising a particular figure of merit. In this study we investigate variance minimising or information-theoretic acquisition functions focusing primarily on exploration rather than exploitation. The first is \textbf{NIPV} (Negative Integrated Posterior Variance), which selects the point that reduces the global posterior variance the most \citep{chen2014nipv}. It is defined as
\begin{equation}
    \text{NIPV} = -\int_X \mathbb{E}[\text{Var}(f(x) \mid D_{\vx}) \mid D] dx,
    \label{eq:nipv}
\end{equation}
where $D_x := D \cup\{\vx_*, y_*\}$.

The second acquisition function tested was \textbf{BALM} (Bayesian Active Learning Mckay) \citep{mackay1992information}, which is an entropy-based acquisition function and queries the point which gives the largest reduction in predictive marginal entropy over possible new locations $\vx$ and parameters $\vtheta$ given $D$. Approximating the posterior over the hyperparameters $p(\theta \mid D)$ gives the following criterion
\begin{align}
    \text{BALM} &= \sH \left [ \mathbb{E}_{p(\theta \mid D)} [p(y \mid \vx, \theta)] \right],
    \label{eq:balm}
\end{align}
and $\vx_*$ is chosen such that the above is maximised.

The final acquisition function was \textbf{BALD} (Bayesian Active Learning by Disagreement), which maximizes the \enquote{disagreement} in predictive posterior entropy and the hyperparameters \citep{houlsby2011bald, kirsch2019batchbald},
\begin{equation}
    \text{BALD} = \sH \left [ \mathbb{E}_{p(\theta \mid D)} [y \mid \vx, D, \theta] \right] - \mathbb{E}_{p(\theta \mid D)} [\sH [y \mid \vx, \theta]].
    \label{eq:bald}
\end{equation}
Here, the entropy is computed in both the parameter and output space to maximize the conditional mutual information gain between the model parameters and output. Intuitively this means one wants to query the point where the uncertainty in the parameters compared to the uncertainty in the output is the largest.

\subsection{Experimental validation}
To evaluate the accuracy of the SDL's prediction of deposition flux distributions, combinatorial CuSn thin films were deposited on amorphous silica substrates, and the composition gradients of the films then characterised by Rutherford Backscattering Spectrometry (RBS). These RBS measurements were performed using the 5-MV NEC-5SDH-2 tandem accelerator at Uppsala University \citep{Strom_2022}, with a primary ion-beam of 4He+, at an energy of 2 MeV. Primary ions had a 0\textdegree~angle of incidence, relative to the sample-surface normal, and the energy of the backscattered ions was measured using a PIP-type silicon detector, placed at an angle of 172\textdegree~relative to the path of the primary ion-beam. Measured RBS spectra were fitted with the SIMNRA software package \citep{MAYER2014176}, using Rutherford scattering cross-sections for all elements, stopping powers derived by DPASS \citep{SCHINNER201919} for all fitted sample-layer compositions, and dual-scattering effects accounted for. The choice to deposit the films on amorphous silica substrates was made to remove the possibility of ion-channelling effects \citep{Nordlund_2016} during RBS measurements, which would compromise the possibility to use signals from the substrate to constrain the particle-fluence during analysis. Measurements were made at 9 positions on each substrate in two perpendicular directions.

\section{Results}\label{sec:results}
In this section we compare the results of the different acquisition functions and choose the optimal approach for our SDL. Then, we combine models trained on single metal sources to predict co-sputtering rates on the QCM sensors and use these to fit the flux model. Finally, we demonstrate an experimental verification by RBS.

\subsection{Comparison of acquisition functions}
In this section, we present results comparing the performance of normal and fully Bayesian GPs initialised using the three different Cases described previously, using NIPV, BALM and BALD query strategies, having random experiment selection as a baseline. Concretely, performance is measured in terms of the rate of reduction of $E_\text{RMS}$ in \eqref{eq:rmse}, and the enhancement factor $F$ in \eqref{eq:ef} relative to random selection, with each queried data point. In other words, high performance corresponds to a smaller number of experiments being needed to train an accurate model.
\begin{figure}
    \centering
    \includegraphics[width=0.5\linewidth]{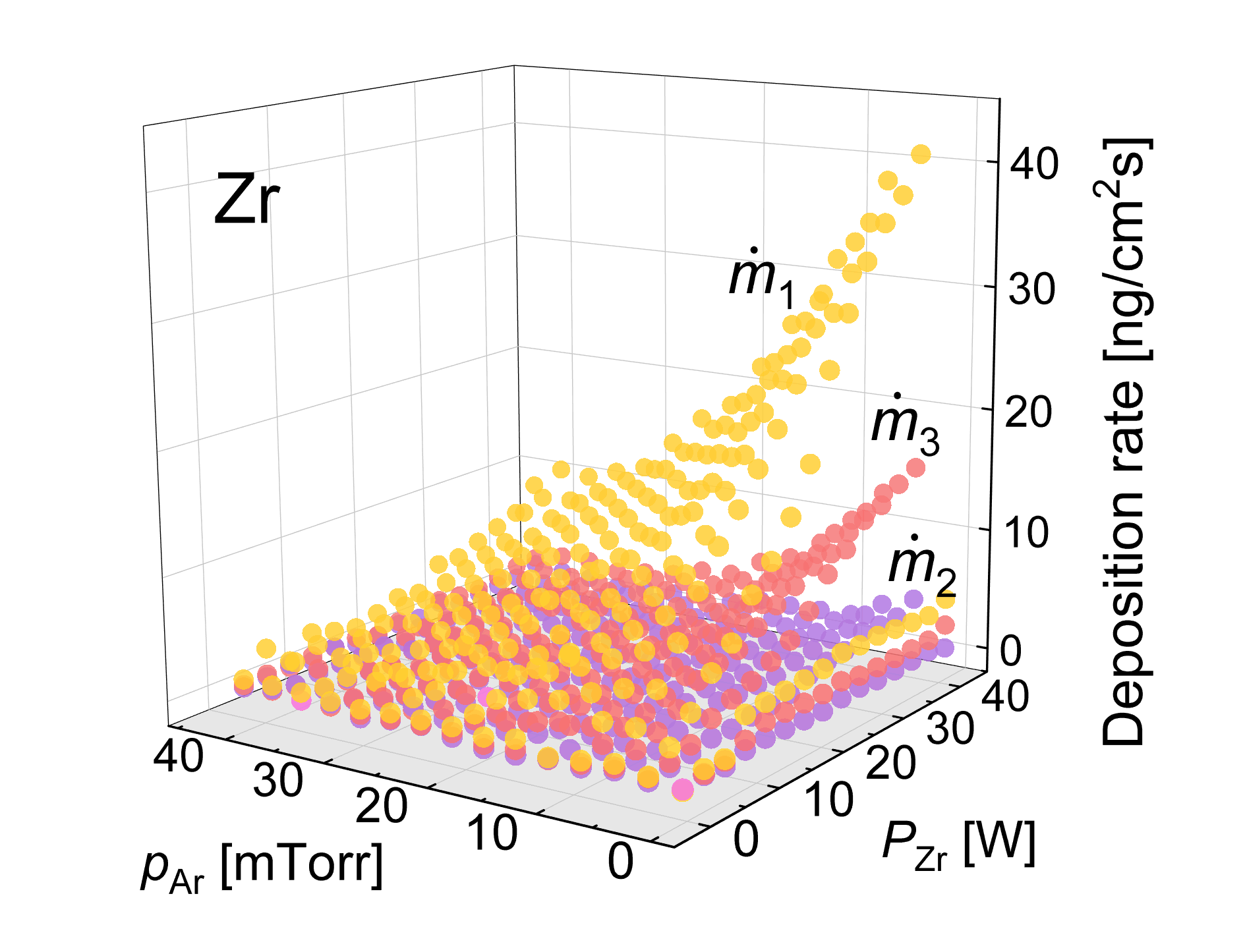}
    \caption{\enquote{Ground truth} dataset of 225 data points collected while sputtering from a Zr target, where $\dot{m}_i$ is the mass deposition rate measured by sensor $i$.}
    \label{fig:zr_grid}
\end{figure}

Figure \ref{fig:zr_grid} shows an example ground truth dataset for sputtering of Zr, obtained by the SDL, that we used to develop the models prior to performing real-time experiments for various other sources. It exhibits typical trends: the deposition rate increases roughly linearly with applied power, while for higher pressure more collisions between atoms occur, which reduces material transport and thus the deposition rate. However if the pressure is too low, the plasma extinguishes and sputtering ceases, hence the sharp drop in deposition rate (close to $p_{Ar} = 0$ in Figure \ref{fig:zr_grid}). Modeling such a sudden drop using GPs requires many samples, as it deviates from the smooth functional forms favoured by the prior. Circumventing this source of inefficiency underlies the choices of initialisation procedures for Cases 2 and 3 described earlier. 

\begin{figure}
    \begin{subfigure}[b]{0.48\textwidth}
        \centering
        \includegraphics[width=\textwidth]{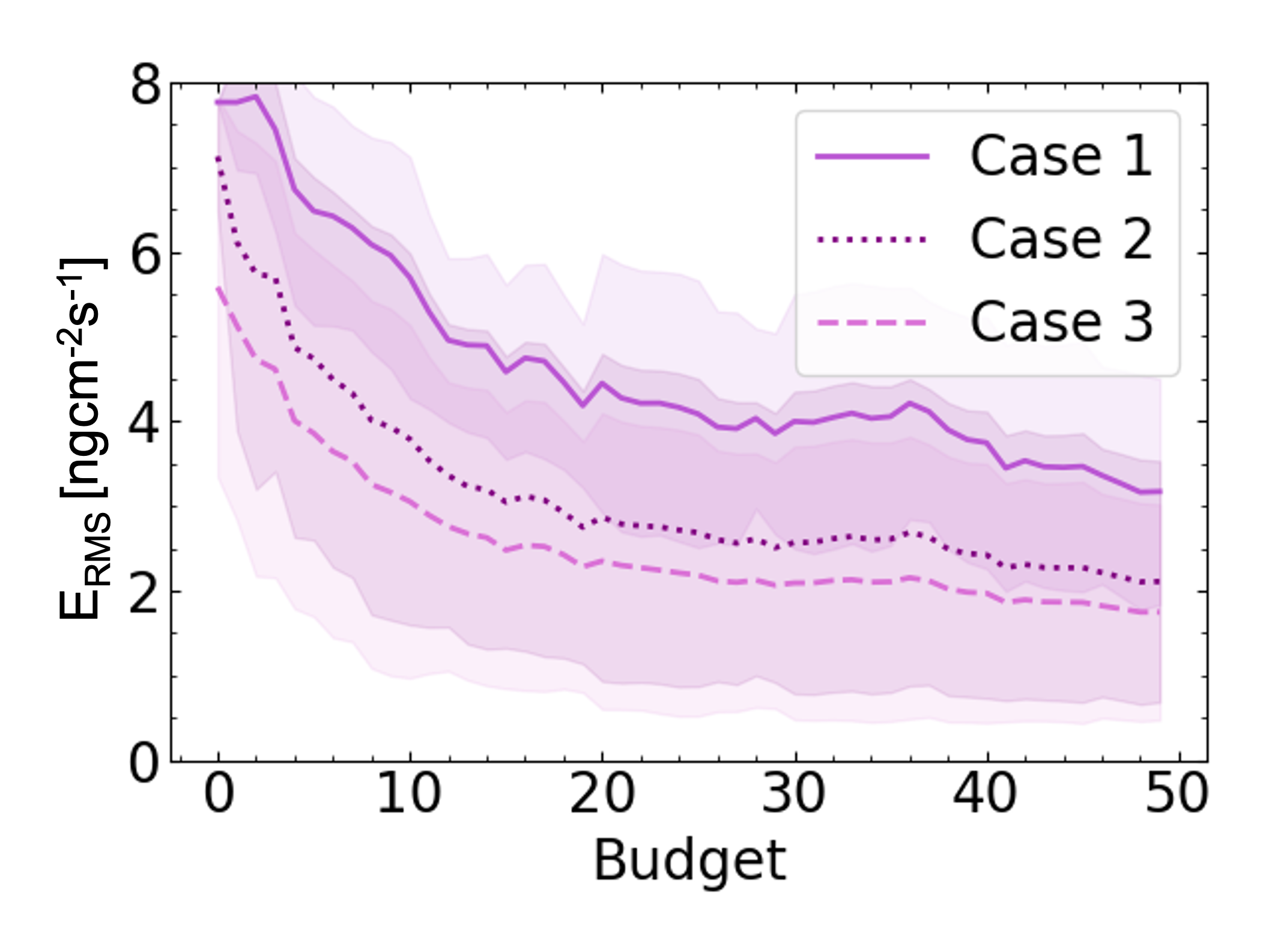}
        \caption{Random}
        \label{fig:random_rmse}
    \end{subfigure}
    \hfill
    \begin{subfigure}[b]{0.48\textwidth}
        \centering
        \includegraphics[width=\textwidth]{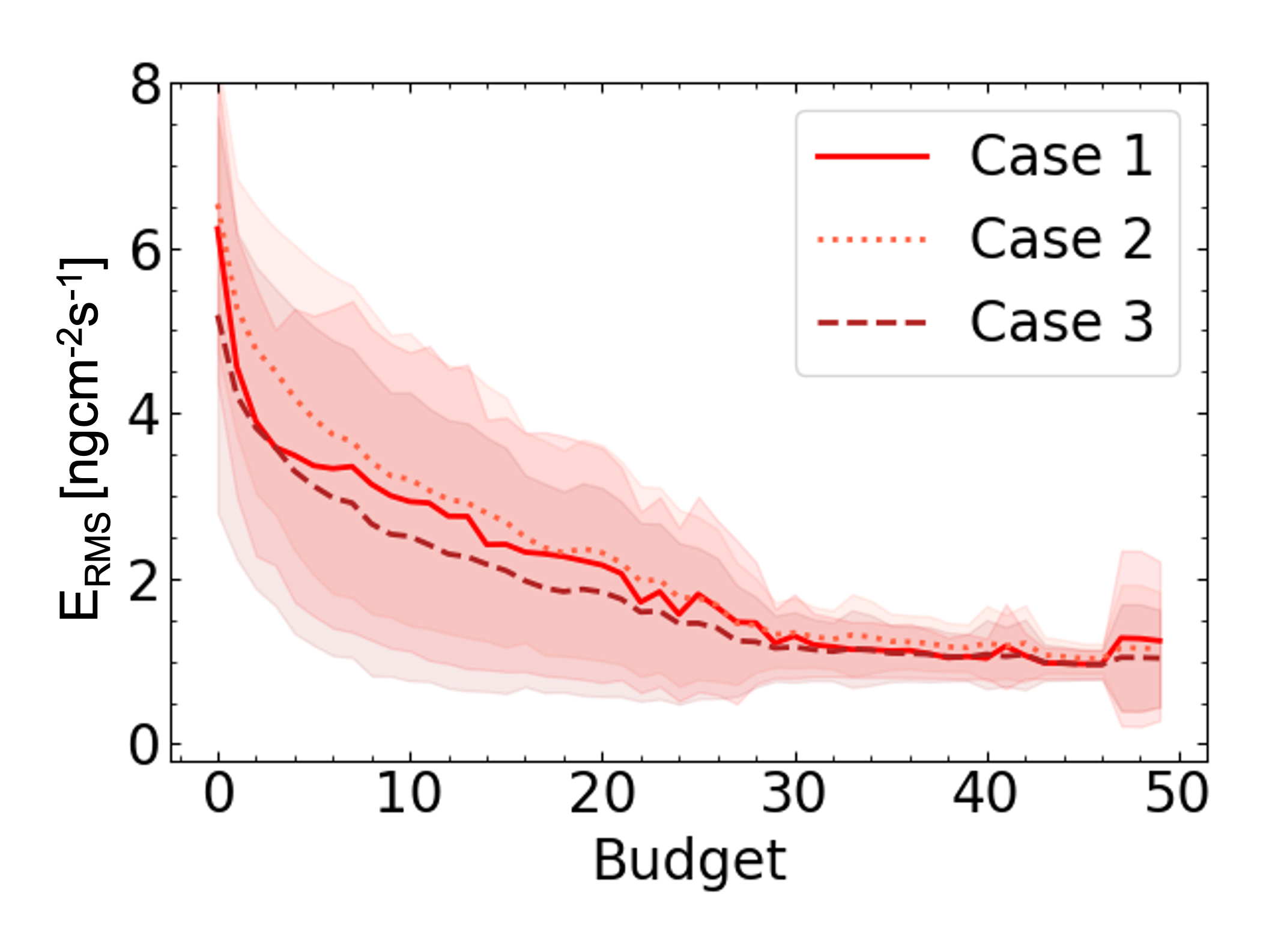}
        \caption{NIPV}
        \label{fig:nipv_rmse}
    \end{subfigure}
    \begin{subfigure}[b]{0.48\textwidth}
        \centering
        \includegraphics[width=\textwidth]{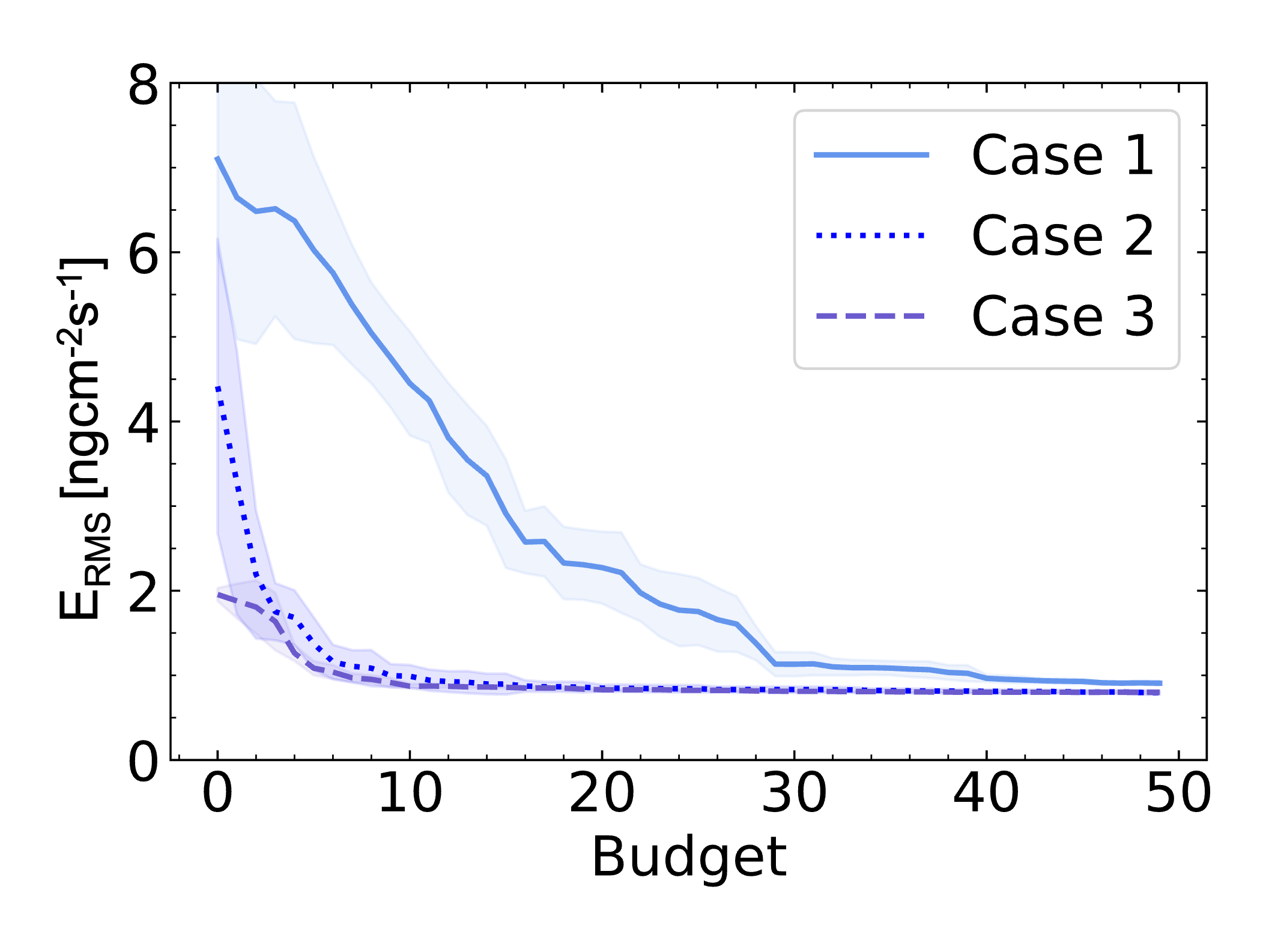}
        \caption{BALM}
        \label{fig:balm_rmse}
    \end{subfigure}
    \hfill
    \begin{subfigure}[b]{0.48\textwidth}
        \centering
        \includegraphics[width=\textwidth]{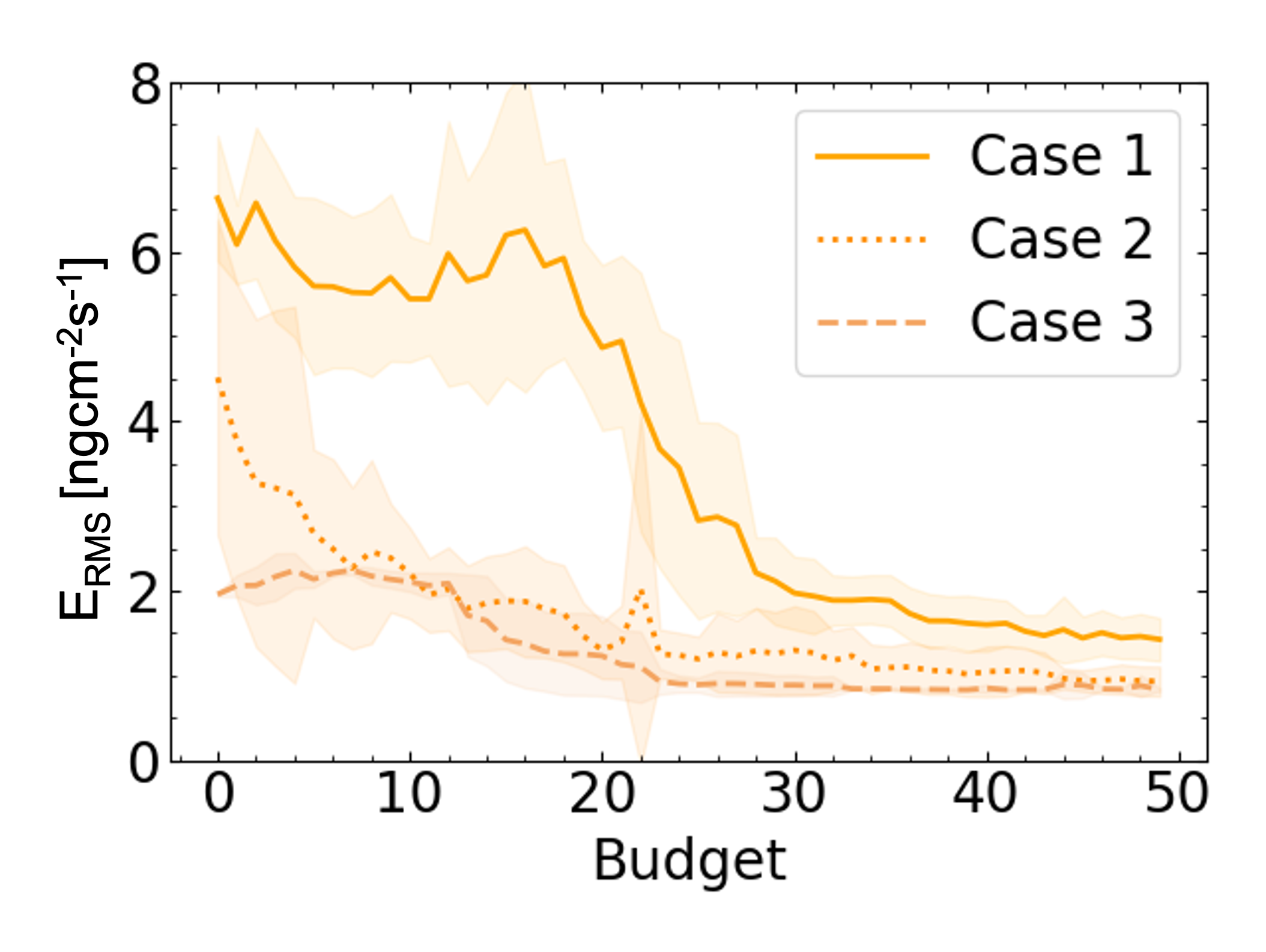}
        \caption{BALD}
        \label{fig:bald_rmse}
    \end{subfigure}
    \caption{Comparison of various acquisition functions running on the Zr dataset in Figure \ref{fig:zr_grid} (a) Random query, (b) NIPV, (c) BALM and (d) BALD, for different choice of initial training points in Cases 1-3. The shaded region represents $\pm 1$ std. We see that BALM in Case 3 achieves the lowest $E_\text{RMS} [ngcm^{-2}s^{-1}]$ score in the fewest queries}
    \label{fig:init_strategies}
\end{figure}

Figure \ref{fig:init_strategies} shows the performance of the three GP initialisation (Cases 1--3) using the different acquisition functions, in terms of $E_\text{RMS}$ defined by \eqref{eq:rmse}, running on the Zr dataset in Figure \ref{fig:zr_grid}. The shaded region around each trace is the standard deviation over 10 runs with 50 queries each. 

In the baseline case using random queries, shown in Figure \ref{fig:random_rmse}, we see that the $E_\text{RMS}$ decreases as more points are acquired until the budget is spent without reaching a stable value for any Case. Case 3 achieves the lowest $E_\text{RMS}$ ($< 2$). The variance is large and fairly constant throughout for all Cases. The large variance arises from random selection of points does not guarantee sampling from all informative parts of the space, leading to larger errors. The same effect is responsible for the steady reduction in improvement rate of $E_\text{RMS}$, which stays well above 0 even after 50 queries.

The general trend for NIPV (Figure \ref{fig:nipv_rmse}) is similar to random, with Case 3 achieving the best performance. Unlike random queries, Cases 2 and 3 do reach a similar, stable, $E_\text{RMS}$ score after about 30 queries. The variance in all Cases becomes significantly smaller after 30 queries. It is expected that NIPV performs significantly better than random, as it deliberately queries points to cover the entire space, hence contributing to faster learning and minimising the global prediction uncertainty. Cases 2 and 3 reach peak performance after a similar number of queries, showing that the initial points are of less importance when using an acquisition function that selects points adaptively. 

For BALM (Figure \ref{fig:balm_rmse}), $E_\text{RMS}$ plateaus around 1 in less than 10 queries for Cases 2 and 3. In contrast, it declines much more slowly in Case 1, reaching the same performance only after 50 queries. Compared to NIPV, there is a larger difference in $E_\text{RMS}$ performance for the three Cases: in particular, cutting off the discontinuity caused by too low chamber pressure (Cases 2 and 3) significantly speeds up the learning process. This behaviour is expected as then no queries need to be spent to learn the sudden drop in deposition rate occurring at low pressures. BALM has lower variance than the other two query methods, as expected from its fully Bayesian model. Unlike standard GPs, fully Bayesian GPs account for hyperparameter uncertainty, making them more robust and less prone to overfitting. By sampling from the hyperparameter prior $p(\vtheta)$, they learn more accurately and achieve lower, more confident $E_\text{RMS}$ scores with fewer queries. This phenomenon highlights the importance of modelling hyperparameter uncertainty, not solely the prediction uncertainty.

The performance of BALD (Figure \ref{fig:bald_rmse}) is similar to BALM, with Cases 2 and 3 achieving much better performance than Case 1. However, BALD needs more than twice as many queries as BALM to reach peak performance: Case 3 reaches an $E_\text{RMS}$ score around 1 in around 25 queries, whereas Case 2 needs around 45 queries to reach the same score. The variance is also higher overall for BALD than BALM. 
\begin{figure}
    \begin{subfigure}[b]{0.48\textwidth}
        \centering
        \includegraphics[width=\textwidth]{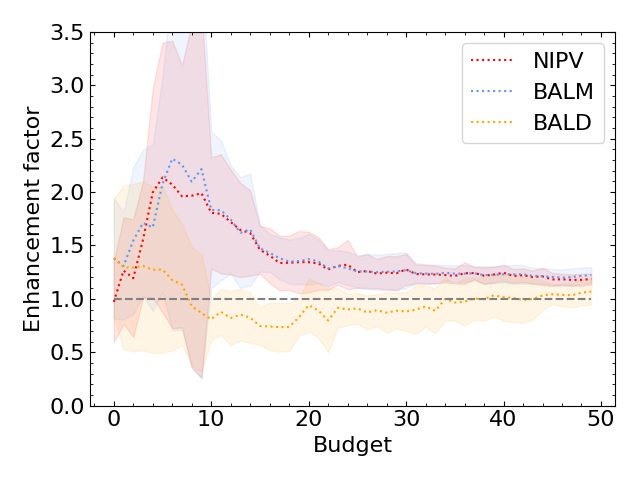}
        \caption{Case 2}
        \label{fig:ef_case2}
    \end{subfigure}
    \hfill
    \begin{subfigure}[b]{0.48\textwidth}
        \centering
        \includegraphics[width=\textwidth]{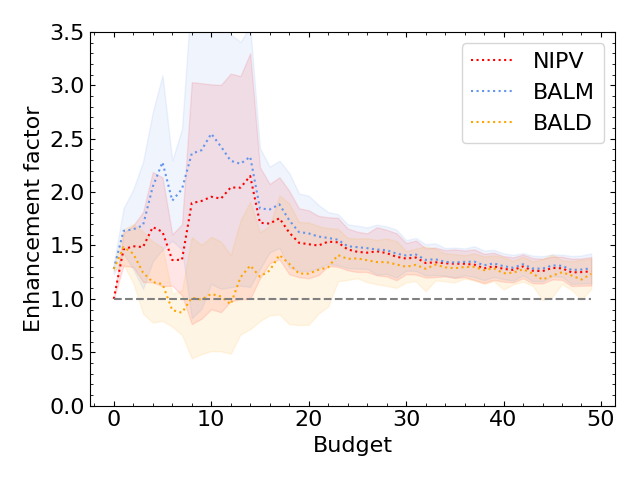}
        \caption{Case 3}
        \label{fig:ef_case3}
    \end{subfigure}
    \caption{The enhancement factor for the different acquisition strategies in relation to random queries (grey dashed line) for (a) Case 2 and (b) Case 3.}
    \label{fig:enhancement_factor}
\end{figure}

Figure \ref{fig:enhancement_factor} shows the enhancement factor defined by \eqref{eq:ef} for the different acquisition functions in Cases 2 and 3, where the grey dashed line is the baseline (random sampling). We see that in Case 2, NIPV and BALM perform on average twice as well as random querying, meaning the $E_\text{RMS}$ of their model fits is half that of random querying, after 10 experiments. For Case 3, all query methods are better compared to random queries, with NIPV and especially BALM producing model fits nearly twice as good as random for the same number of queries. BALD is also better than random, compared to Case 2. Again, this stems back to the fact that fully Bayesian GPs are more robust, and it demonstrates the importance that the choice of initial points has for accurate model fitting; if it is possible to make a physically-relevant and unbiased choice of initialisation points, then this can be very advantageous. It is evident that BALM outperforms the others, obtaining the lowest $E_\text{RMS}$ in the fewest queries $< 10$ with the lowest variance.
\begin{figure}
    \begin{subfigure}[b]{0.48\textwidth}
        \centering
        \includegraphics[width=\textwidth]{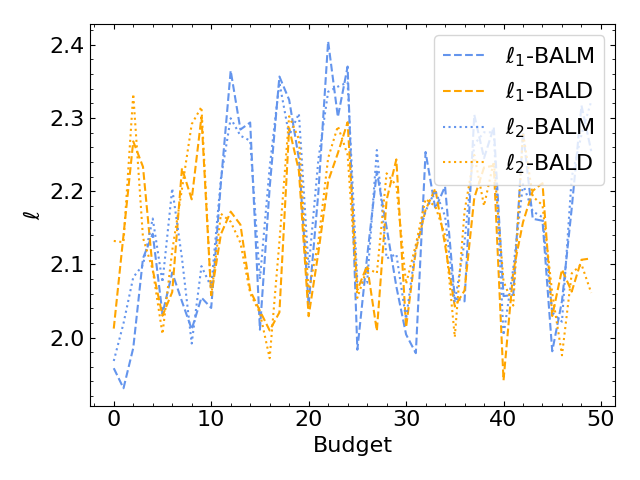}
        \caption{}
        \label{fig:ls_mean}
    \end{subfigure}
    \hfill
    \begin{subfigure}[b]{0.48\textwidth}
        \centering
        \includegraphics[width=\textwidth]{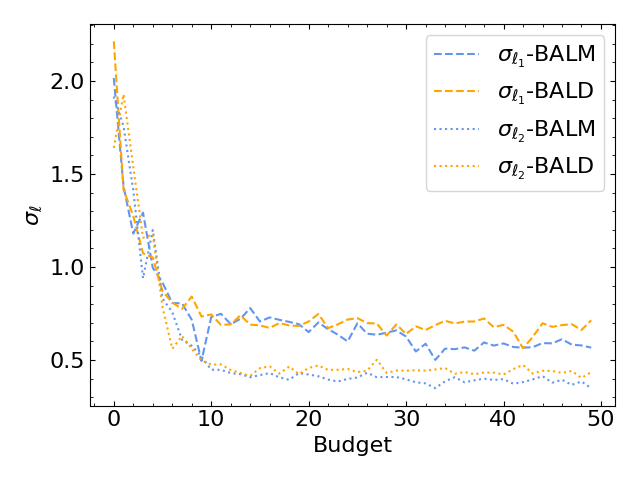}
        \caption{}
        \label{fig:ls_std}
    \end{subfigure}
    \caption{(a) The median lengthscales in both input dimensions averaged over the 10 runs. (b) An example of the standard deviation of the lengthscales $\ell_1, \ell_2$, for the 1024 MC samples at each iteration.}
    \label{fig:hyperparams}
\end{figure}

To understand the different behaviour between BALM and BALD, we compare the evolution of the hyperparameter fits as more measurements are acquired. Figure \ref{fig:hyperparams} shows the median lengthscales in both input dimensions $\{\ell, \sigma^2_n\}$ averaged over 10 runs. BALM and BALD behave similarly, with a largely constant median lengthscale and quickly concentrating distributions in the first 10 iterations. The small difference excludes the hyperparameters from being the explanation for the different sample efficiency between BALM and BALD. which rather results from the inherent difference of the acquisition functions.

For application in the SDL, the above results make it clear that, regardless of acquisition function, restriction of the power-pressure space to avoid the plasma extinguishing (Case 2) is critical for rapid learning of $\dot{m}_i$. The precise cut-off pressure will depend on material and applied power, and could potentially be learned by a separate algorithm; hereafter we simply let the minimum pressure be 4 mTorr for all cases. Making an informed choice of initialisation points (Case 3) is useful but not critical, for reasonable acquisition functions. Based on these results and that different sources follow a similar behaviour, we decided to implement Case 3 in practice in the SDL. 

In the SDL application, the overall cycle time is important, which includes both the experiment time, about 30 seconds per point, and the computation time, which is variable depending on computing power and choice of acquisition function. In this case, running on a desktop CPU, the computation time for BALM and BALD was on the order of minutes, compared to seconds for NIPV. Thus, choosing NIPV is the most time-efficient. However, if the compute power were to be increased, BALM is the best option. Ultimately, both choices allow model training for a given target in only about 15 minutes, which is comparable to the time taken for a typical thin film deposition. This hints at the possibility of an asynchronous workflow where the next measurement is computed during sputtering.

\begin{figure}
    \begin{subfigure}{0.33\textwidth}
        \centering
        \includegraphics[width=\textwidth]{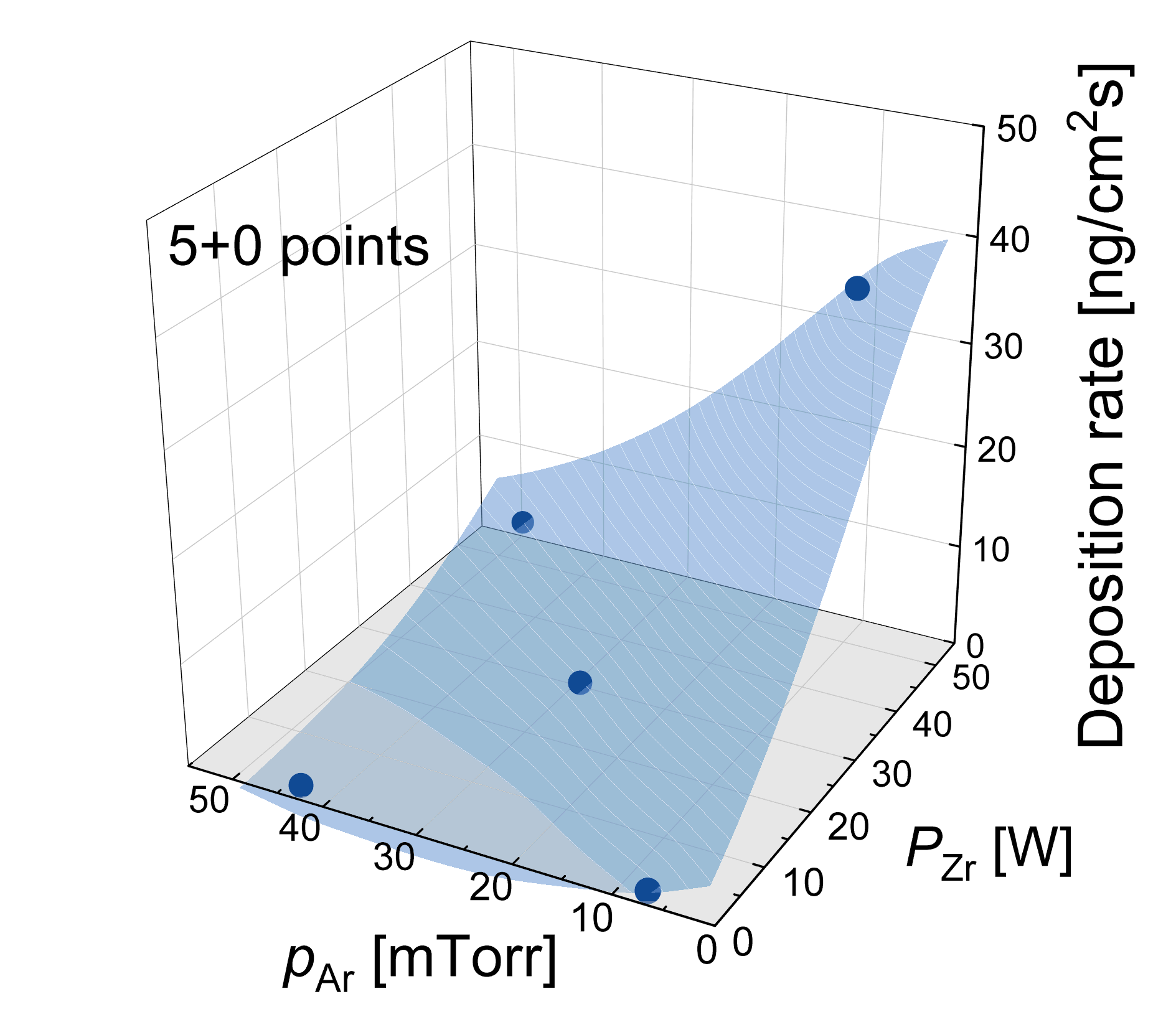}
    \end{subfigure}
    \hspace{-5mm}
    \begin{subfigure}{0.33\textwidth}
        \centering
        \includegraphics[width=\textwidth]{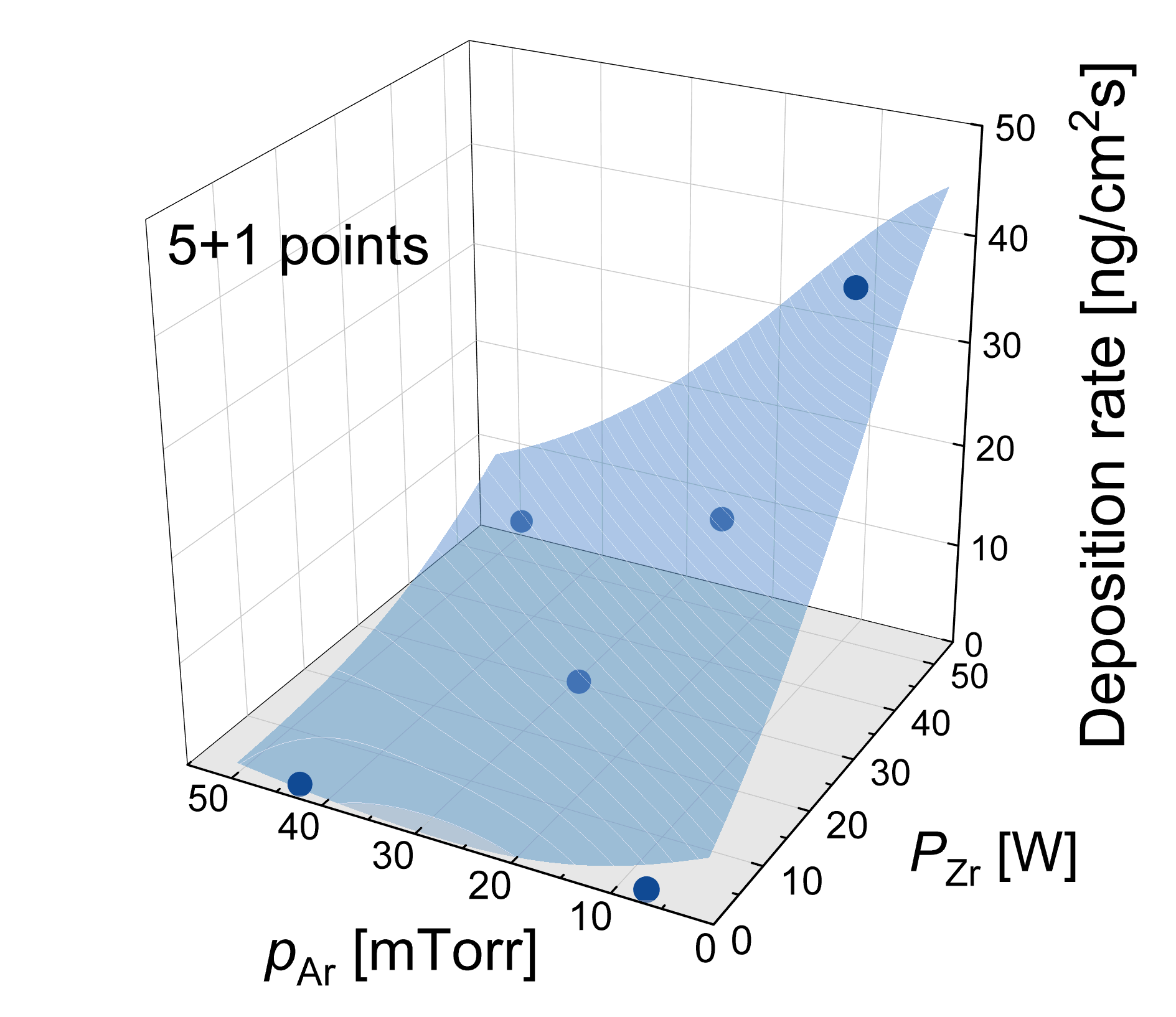}
    \end{subfigure}
    \hspace{-5mm}
    \begin{subfigure}{0.33\textwidth}
        \centering
        \includegraphics[width=\textwidth]{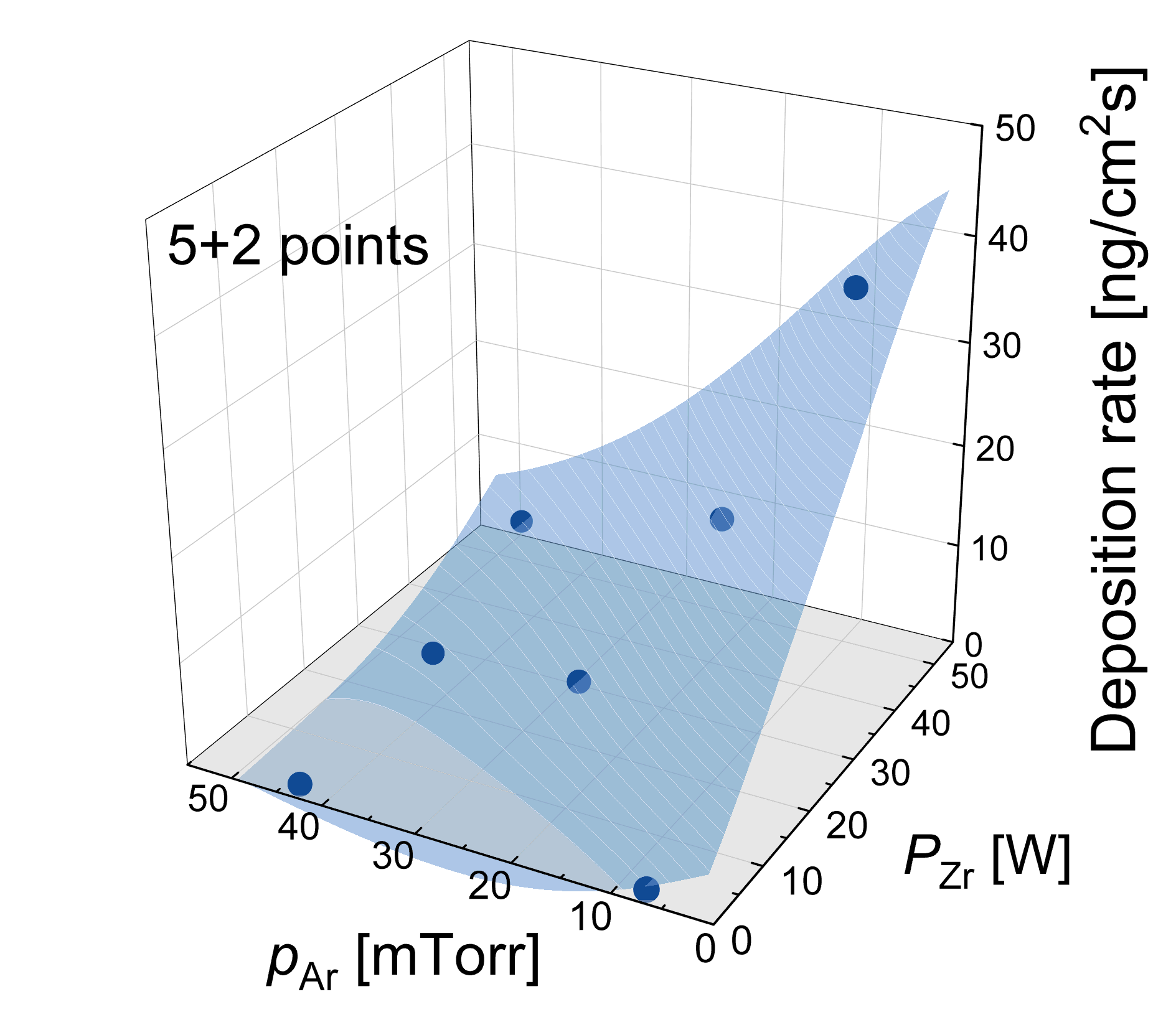}
    \end{subfigure}
    \begin{subfigure}{0.33\textwidth}
        \centering
        \includegraphics[width=\textwidth]{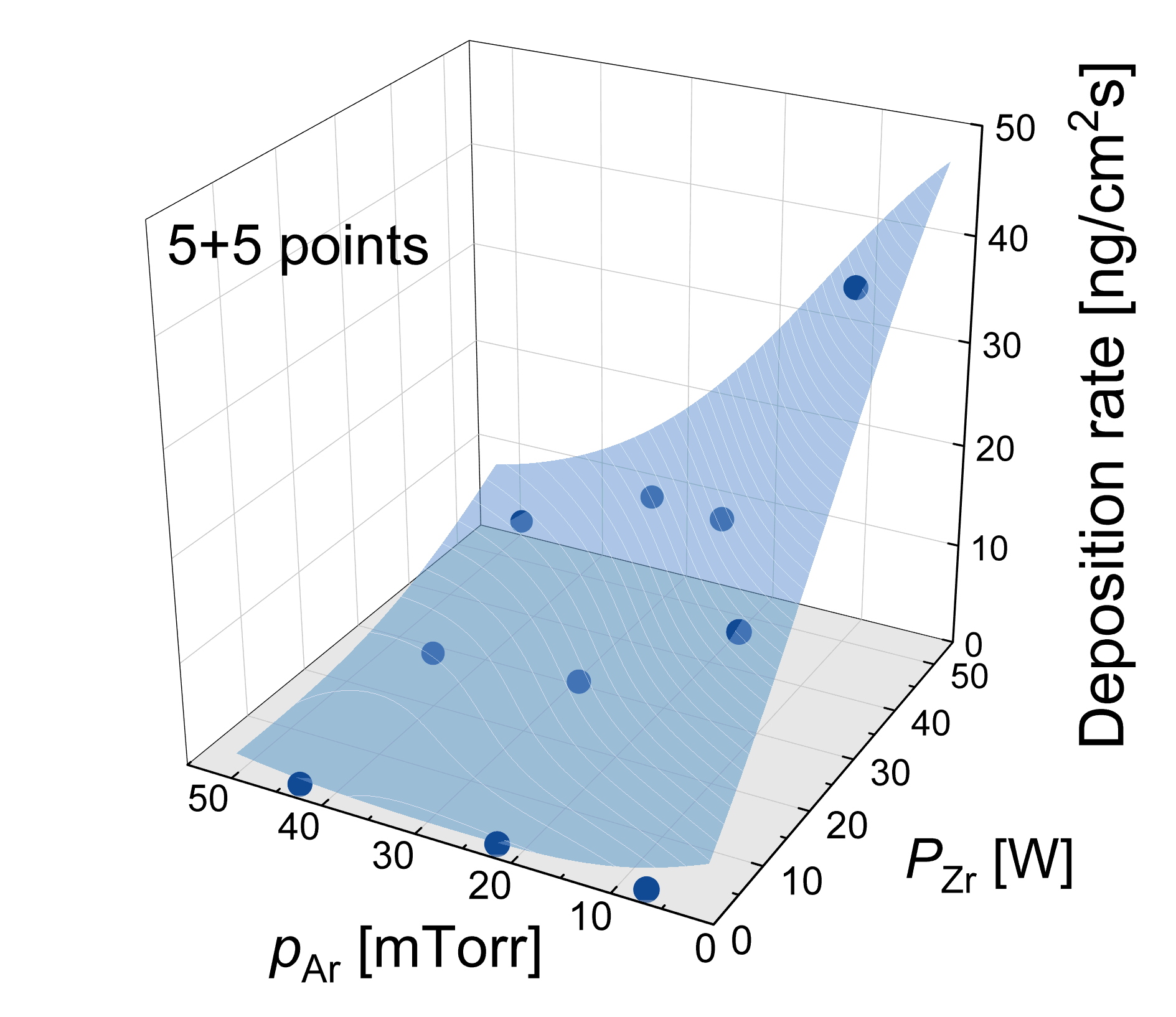}
    \end{subfigure}
    \hspace{-2mm}
    \begin{subfigure}{0.33\textwidth}
        \centering
        \includegraphics[width=\textwidth]{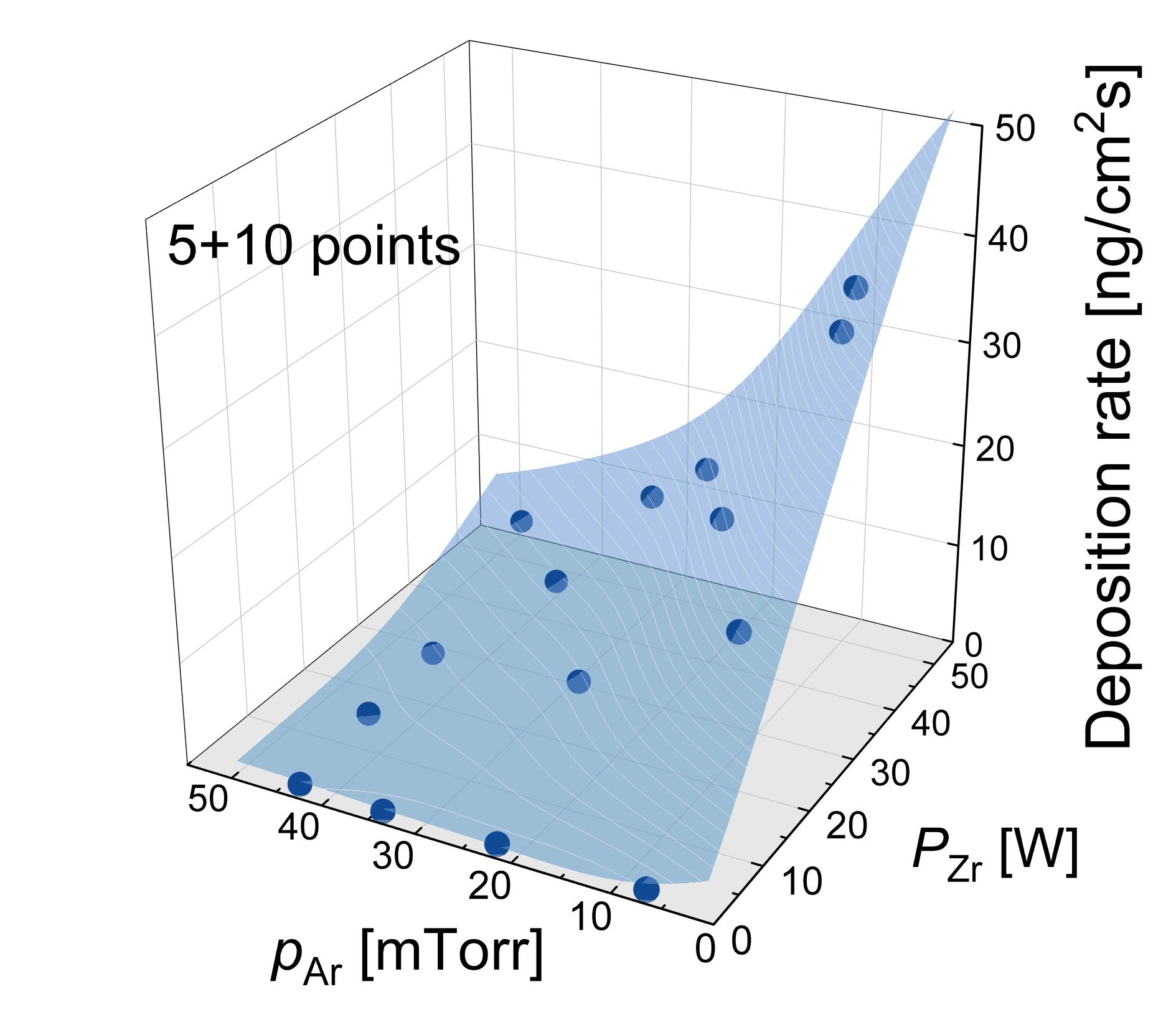}
    \end{subfigure}
    \hspace{-2mm}
    \begin{subfigure}{0.33\textwidth}
        \centering
        \includegraphics[width=\textwidth]{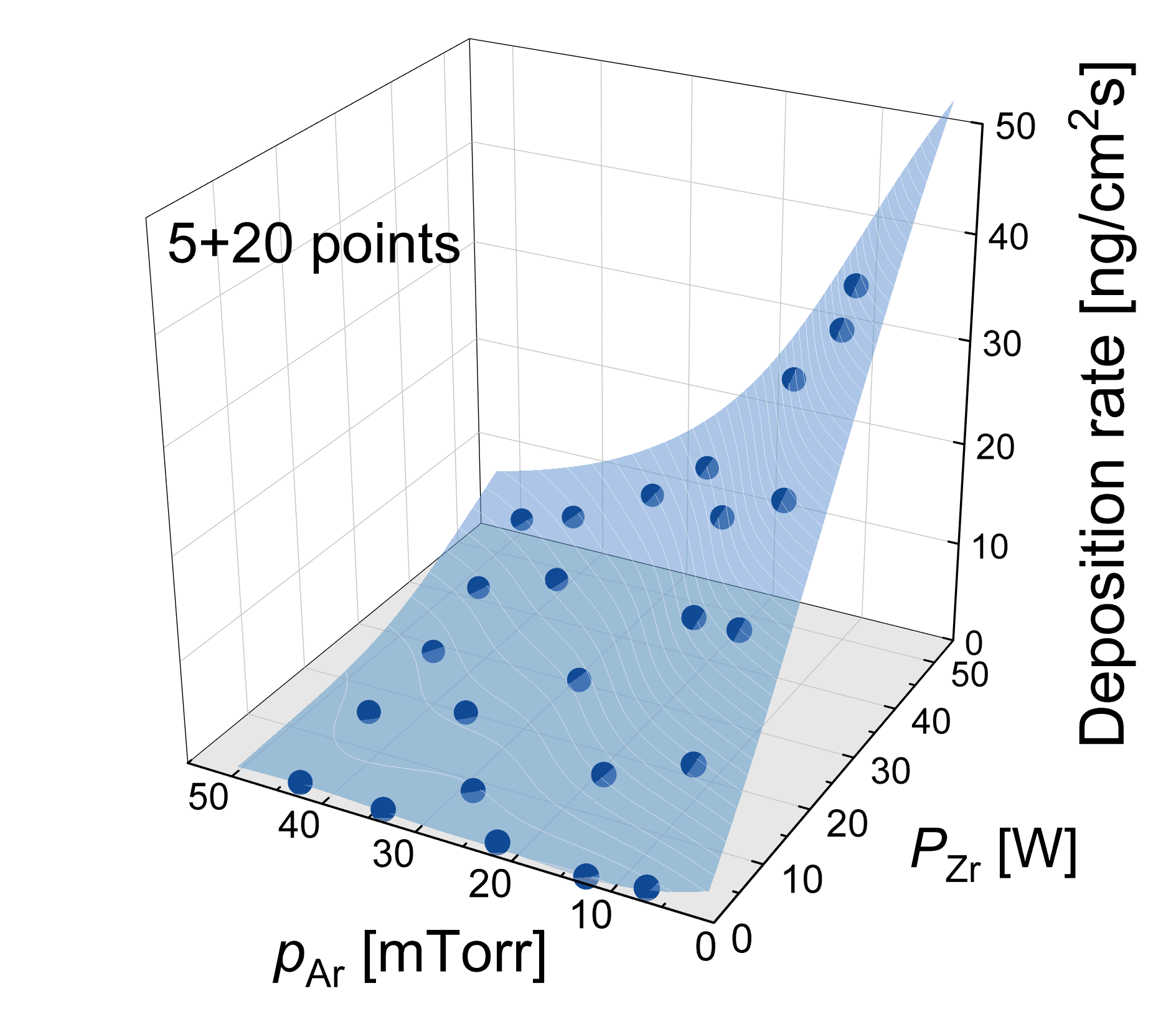}
    \end{subfigure}
    \caption{Evolution of a GP using BALM acquisition strategy, in Case 3 after 0, 1, 2, 5, 10 and 20 queries. Once 10 new points have been acquired, the GP manages to create a good fit of the data points.}
    \label{fig:gp_evolution}
\end{figure}

Figure \ref{fig:gp_evolution} illustrates the iterative process of fitting a Fully Bayesian GP with BALM, initialised with 5 points according to Case 3, and showing the predicted deposition rates after up to 20 new queries. We notice that in the earlier stages, the GP predicts negative (i.e. non-physical) deposition rates, but this is resolved after about 10 new points have been queried. Subsequently, the GP manages to model the entire deposition rate dataset efficiently, with a uniform surface passing through all data points, notably without the creation of localised extrema which would indicate poor fitting or overfitting of the data.  

\subsection{Application and validation for co-sputtering}
Finally we describe application of the developed ML methodologies in our SDL. This entails that datapoints queried by the GP models are obtained via experiment, as opposed to being picked from a pre-existing dataset (as was the case for model development). We use the example of Cu-Sn co-sputtering as an example. The aim for the SDL is to determine experimental conditions ($P_{Cu}$, $P_{Sn}$, $p_{Ar}$) that will generate deposits with a specified composition at the centre of the substrate, a desired thickness, as well as to provide an accurate map of the composition gradients across the substrate.

\begin{figure}
    \begin{subfigure}{0.33\textwidth}
        \centering
        \includegraphics[width=\textwidth]{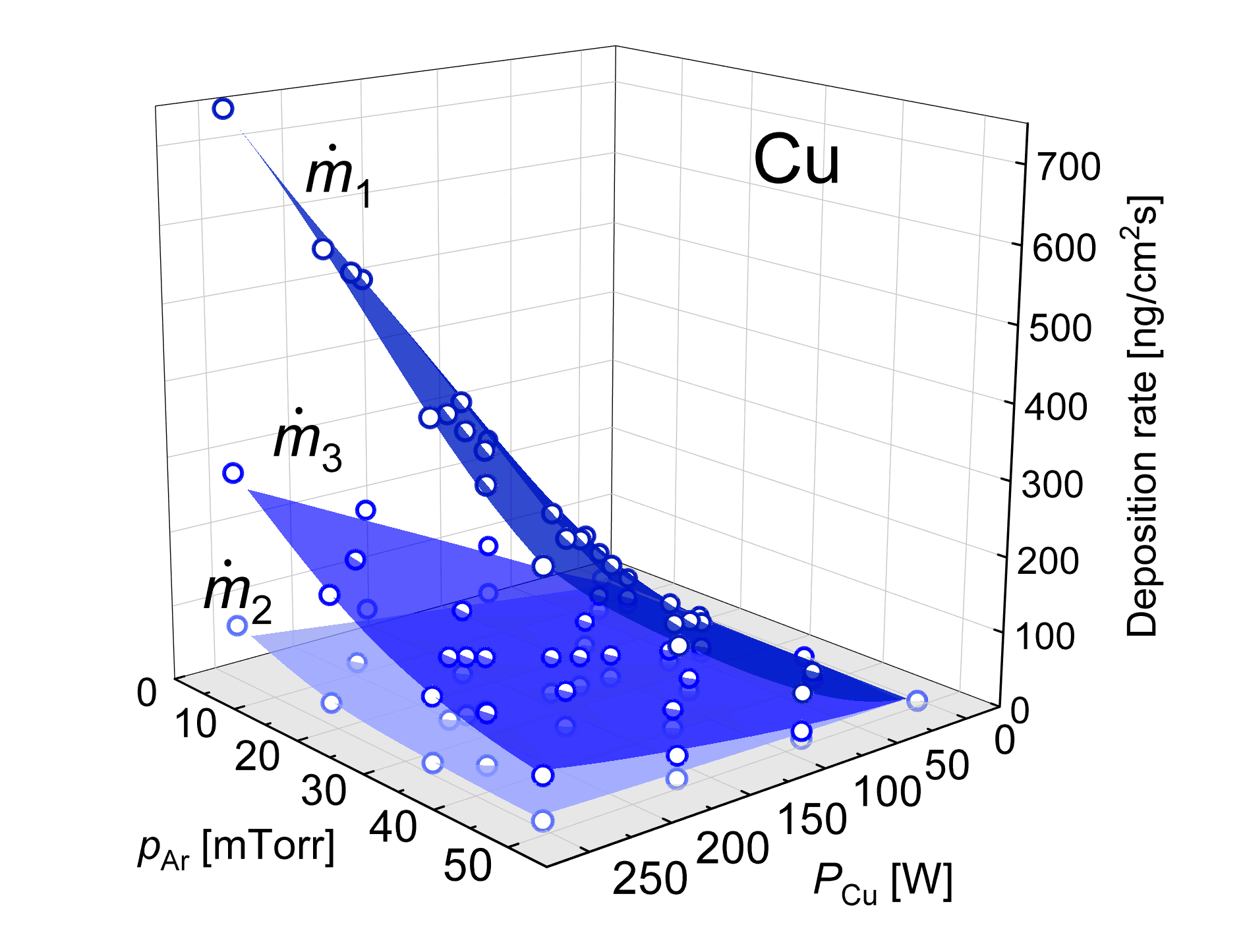}
        \caption{}
        \label{subfig:cu_rate}
    \end{subfigure}
    \hfill
    \begin{subfigure}{0.33\textwidth}
        \centering
        \includegraphics[width=\textwidth]{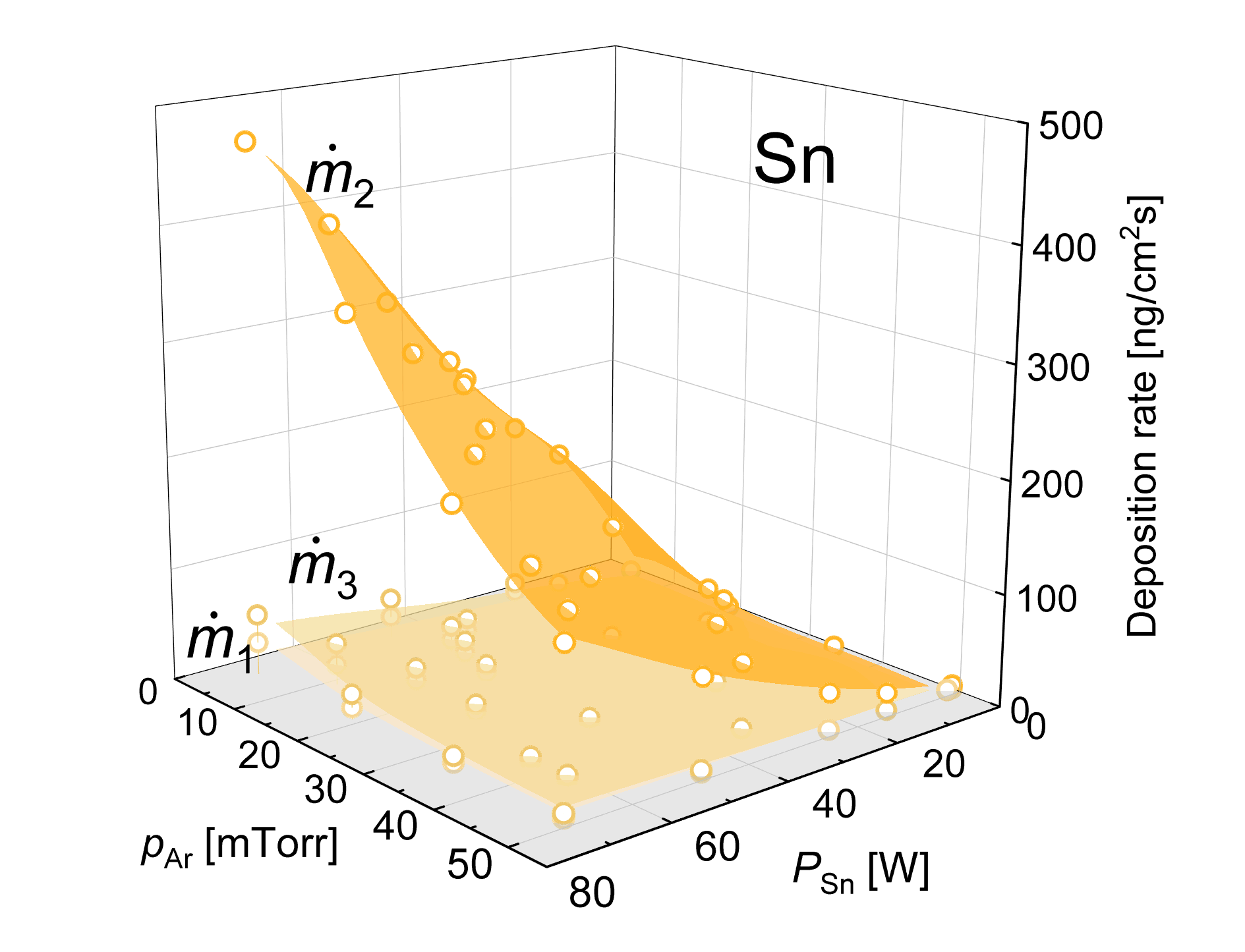}
        \caption{}
        \label{subfig:sn_rate}
    \end{subfigure}
    \hfill
    \begin{subfigure}{0.30\textwidth}
        \centering
        \includegraphics[width=\textwidth]{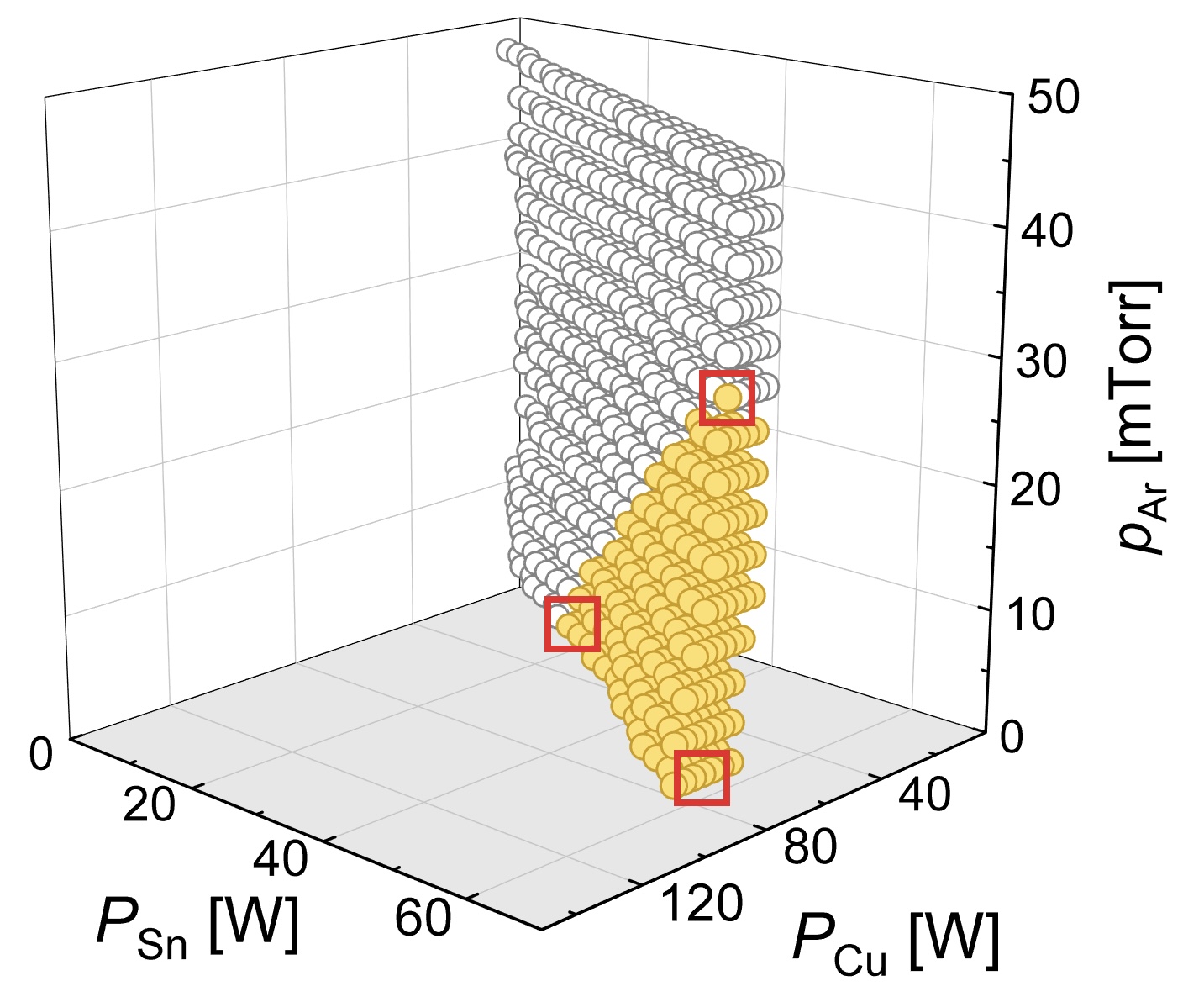}
        \caption{}
        \label{subfig:co_sputter_recipes}
    \end{subfigure}
    \caption{Mass deposition rates at each QCM sensor predicted by GPs with NIPV after 30 iterations, also showing the queried points, for (a) a Cu target and (b) a Sn target. (c) All process parameter combinations yielding a target composition of Cu/(Cu+Sn) = 2 at the substrate center. The yellow points indicate conditions for which a film thickness of 150 nm can be obtained in $\leq 15$ minutes. The highlighted points in red are those used to prepare samples for validation in Figure \ref{fig:RBS}.}
    \label{fig:SDL_model_training}
\end{figure}

The starting point is for the SDL to learn the QCM deposition rates $\dot{m}_{1,2,3}$ for Cu and Sn targets individually. For the practical reasons described, we use the NIPV acquisition function and the \enquote{Case 3} setup, with 30 iterations. Results of GP training are shown in Figure \ref{subfig:cu_rate}, for Sn and Figure \ref{subfig:sn_rate}, for Cu. As expected, the trained GPs provide excellent fits to the measured data. The different positions of the targets with respect to the QCMs account for the trends of deposition rate between the three sensors, with the nearest sensor receiving highest flux. The Sn target has a smaller range of applied power due to the lower melting point of the material.

Moving from single-target sputtering to a co-sputtering situation, a few factors must be borne in mind. First, while $P_{Cu}$ and $P_{Sn}$ are independent, the $p_{Ar}$ is now a common parameter in co-sputtering. Second, the QCM sensor readings measure total mass without distinguishing the Cu and Sn contributions. Finally, it is possible that the co-sputtered elements influence each other's deposition rates, due to chemical interactions or resputtering effects. The SDL can readily test for such effects by performing a set of co-sputtering runs with wide-ranging power and pressure conditions, and comparing the actual QCM mass-deposition rates with the sum of the predicted deposition rates from the individually-trained source GPs. In the absence of interactions, which is not unusual for metal sputtering at room temperature, the sum of predicted deposition rates will be equal to the actual measured rates. This is exactly what can be seen in the example data, for Cu and Sn co-sputtering, presented in the Supplementary Information in Figure S2.

Having learned the deposition rates at the QCM positions for each source, the SDL is requested to derive sputtering conditions that provide a specific composition at the centre of the substrate. To achieve this, it applies the geometrical sputter flux model described in Section \ref{sec:flux_model}, which enables interpolation of deposition rate from the QCM array to all $(x,y)$ locations across the substrate surface. This is initially done for all $P_k$, $p_{Ar}$ combinations, and by converting from mass- to molar-deposition rates, we derive a composition map prediction for any given combination of process conditions. By assuming bulk densities of the constituent elements and integrating over a particular deposition time, one can also predict a thickness map of the deposited film. Subsequently, the subset of conditions that result in a desired centre-point composition and film thickness can be identified. An example is shown in Figure \ref{subfig:co_sputter_recipes} for a targeted composition Cu/(Cu+Sn) = 0.66. As a result of these steps the SDL has narrowed a process parameter space containing, in this example, 875 000 possible sputtering recipes, to a few hundred recipes that provide a desired composition and film thickness. Notably, all of this was achieved automatically in the SDL based on a total of 30 minutes sputtering time (15 minutes to train GPs for each target) without producing a single physical sample.
\begin{figure}
    \begin{subfigure}{0.5\textwidth}
        \centering
        \includegraphics[width=\textwidth]{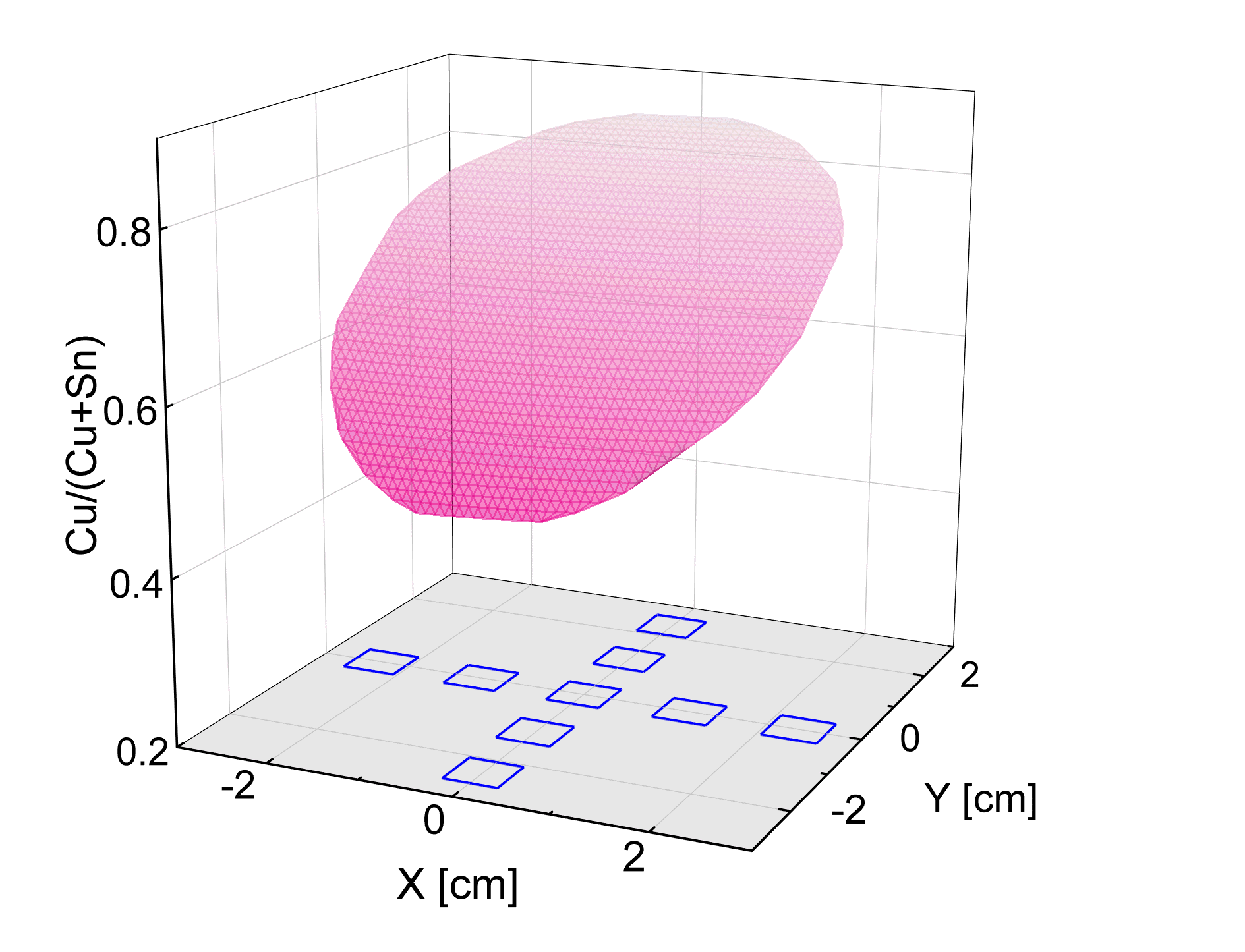}
        \caption{}
        \label{fig:rbs_crossection}
    \end{subfigure}
    \hspace{-2mm}
    \begin{subfigure}{0.5\textwidth}
        \centering
        \includegraphics[width=\textwidth]{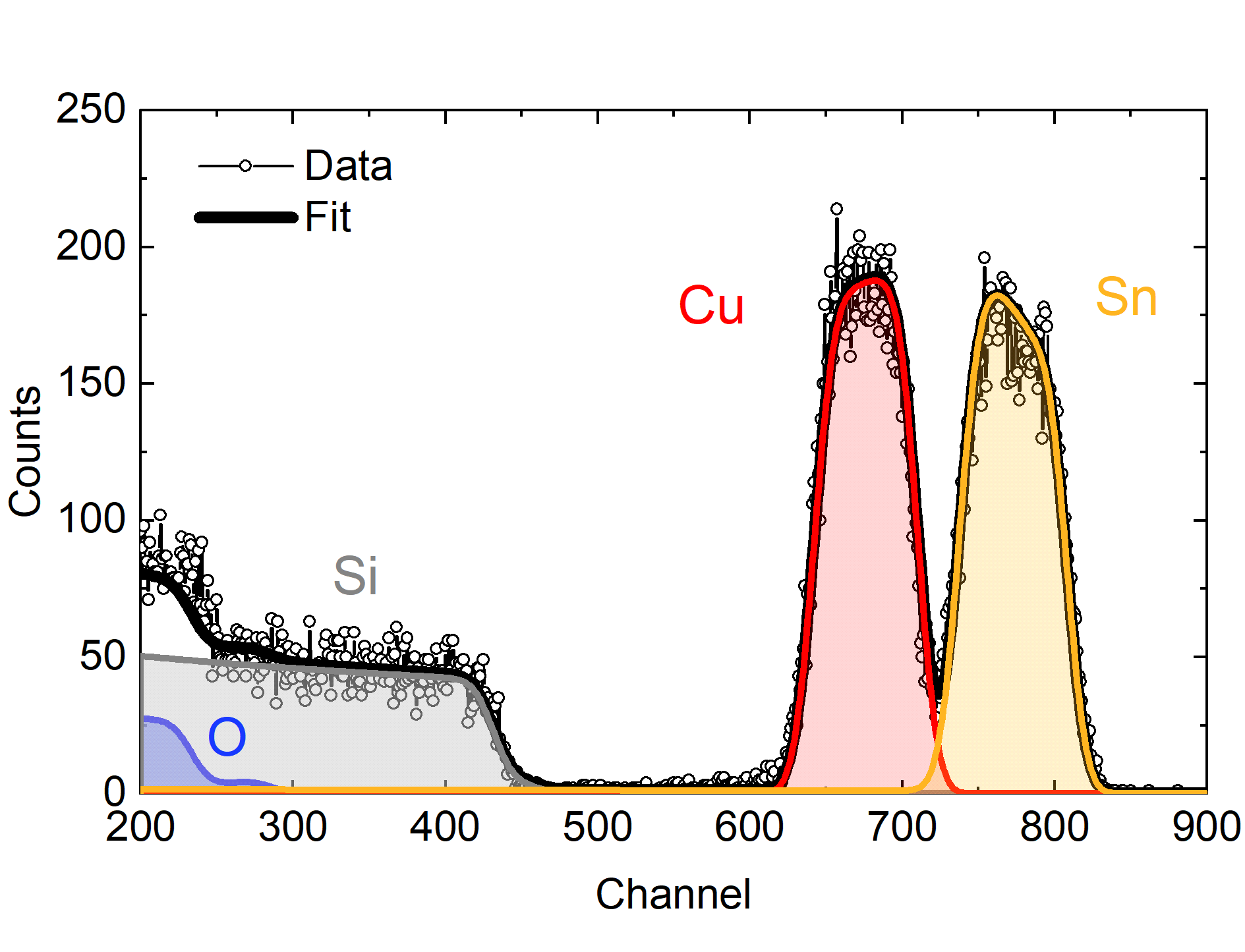}
        \caption{}
        \label{fig:rbs_diagram}
    \end{subfigure}
    \caption{(a) Example composition map predicted for a co-sputtering process, expressed as ratio of Cu/(Cu+Sn) atomic densities, over a 50mm diameter wafer, showing the locations for RBS measurements on the base plane. (b) Example of RBS measurement and fitting at one measured location on a co-sputtered Cu-Sn sample.}
    \label{fig:RBS}
\end{figure}

\begin{figure}
    \centering
    \includegraphics[width=\linewidth]{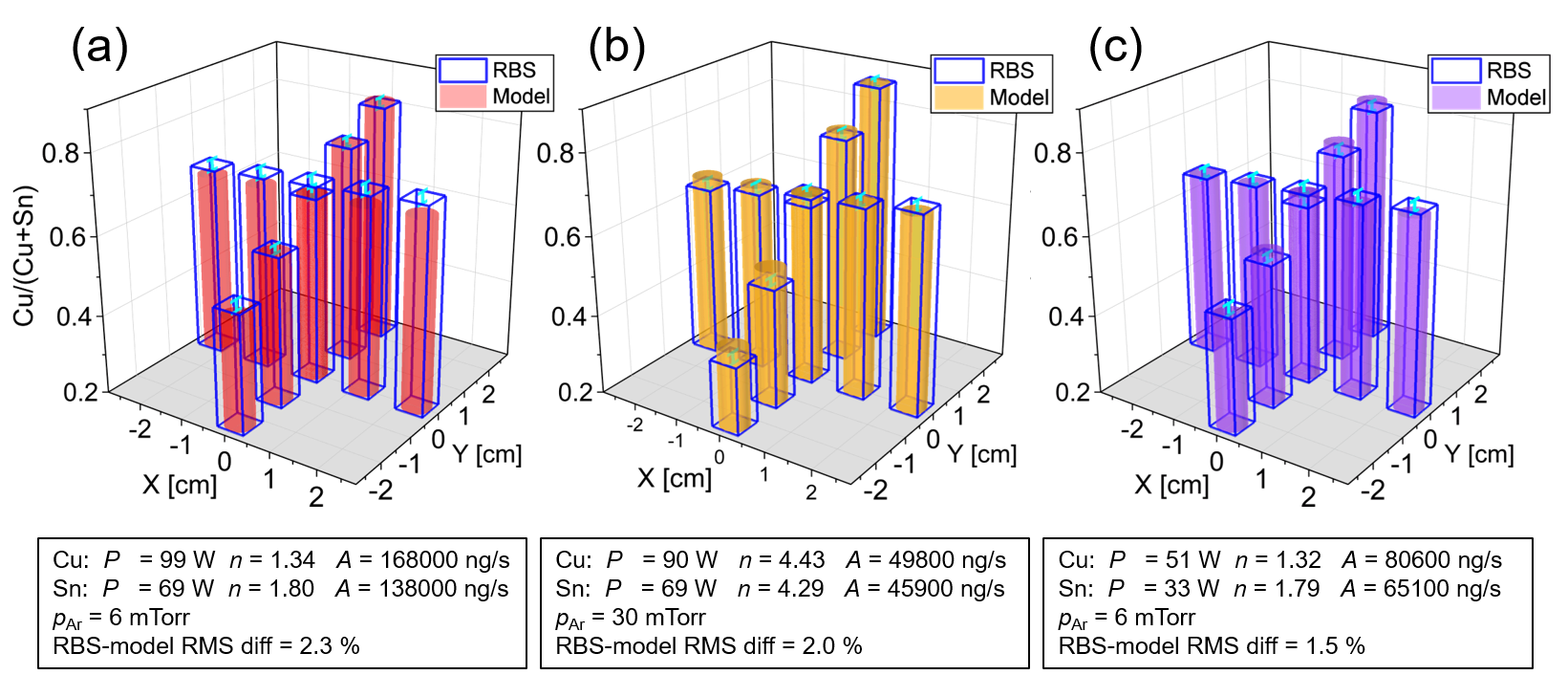}
    \caption{Comparison of modelled (solid bars) and measured (wireframe bars) composition at 9 positions across co-sputtered Cu-Sn samples made with different process conditions, stated in each figure.}
    \label{fig:Cu-Sn-maps}
\end{figure}
To validate the approach, we produce thin film Cu-Sn samples in the SDL, in this instance using three different co-sputtering conditions as indicated by the highlighted points in Figure \ref{subfig:co_sputter_recipes}.  Each condition should provide a centre-point composition Cu/(Cu+Sn) = 0.66 but have different composition spread, due to the different power and pressure combinations used.  An example composition map prediction is shown in Figure \ref{fig:rbs_crossection}. RBS analysis was performed at 9 points across each sample, which were deposited onto 5 cm diameter fused silica wafers, as indicated in Figure \ref{fig:rbs_crossection}. Fitting of the data from each point, exemplified in Figure \ref{fig:rbs_diagram}, provides quantitative composition measurements allowing characterisation of the composition distribution over the wafer. The RBS results are finally compared to the predicted composition maps for each sample in Figure \ref{fig:Cu-Sn-maps}. The agreement between the predictions and the measurements is remarkable, even as the composition spread varies subtly in response to the power and pressure conditions. Root mean square deviations between the modelled and measured points in each 9-point composition map were  2.3\%, 2\% and 1.5\% respectively for the examples shown in Figure \ref{fig:Cu-Sn-maps}. The centre-point compositions were 0.67. 0.65 and 0.65, compared to the targeted value of 0.66.   These deviations are within the error of the  RBS measurement - with the most significant error being the positioning of the sample.

\section{Discussion}\label{sec:conclusions}

By coupling process-specific knowledge and well-chosen active learning approaches, our SDL achieves an extremely powerful capability for high-throughput thin film science, namely accurate prediction of composition maps for co-deposition from arbitrary combinations of sources, without relying on any external calibration. Moreover, we obtain absolute atomic quantities, enabling also thickness maps to be estimated. These results improve greatly upon the relative compositions values and lack of thickness information obtained with standard ex-situ measurements like EDS or XRF mapping. Crucially, the SDL is able to deduce composition maps for \textit{any} possible sputtering process after being trained on the individual targets (15 minutes per target), whereas a single XRF map at low resolution, representing only one sputtering condition, would take at least this long to obtain and would most likely be less reliable than our approach. 

The capabilities demonstrated here will be very useful for materials exploration. Whether the sputtered films are used as-is, e.g. in the context of studying alloy properties, or used as precursors for additional processing stages -- which is the intention in our SDL, using thermal processing in various reactive gases -- the initial compositional and thickness information is critical for deriving synthesis-property relationships. Meanwhile, our approach makes it possible to easily define the full range of co-sputter process to produce a desired composition. Samples from the higher power/lower pressure regimes will have greater density and compressive stress, and vice versa. Thus, one can readily adjust these properties without needing to recalibrate the process to maintain the correct composition. This is important for developing processes that avoid stress-cracking or delamination, that are frequently-encountered problems in thin film science.
 
Naturally, the generality of these results needs to be established for a larger number of target combinations. We have tested two and three-element combinations of Cu, Sn, Zr, Ba, Ti and Ge targets, covering a large range of electronegativities and atomic masses. Some further examples of the linear additivity of multiple sources are shown in the Supplementary information Figure S2 (two sources)  and Figure S3 (three sources).  The assumption of source additivity was only found to be invalid when the sources were adjacent to each other, Ba-Sn in Figure S2. This is presumed to be due to an interaction between their plasmas, detectable as a change in the source voltage, which enhances the sputtering rate compared to single-source sputtering. However, since co-sputtering from closely-placed sources also leads to a narrower composition distribution on the substrate, this situation is less useful for materials exploration and avoiding it comes at little cost. Another likely source of problems will be elements with poor sticking coefficients (high vapour pressures), e.g. Zn. Such elements may have differing adhesion at the substrate compared to the QCM during co-sputtering, due to element interactions, or if substrate heating is used. Figure S2 presents additional comparisons between predicted composition and RBS measurements for three Cu-Ba samples. The agreement is again remarkable in two cases. In the third case - for a high sputter pressure - the match is poorer, while the \textit{n} value for Ba also becomes extreme. This indicates a less physically-realistic solution of the flux model, i.e. a highly directional flux, that is possibly connected to poor sticking of Ba at higher pressures. High values of \textit{n} may therefore be used as an indicator of irregular behaviour. Finally, heavy elements could cause resputtering of co-sputtered elements in some circumstances. This would be harder to notice using QCM data alone and would rely on ex-situ measurements of prepared samples. Extending the approach to reactive sputtering is another future challenge, as the mass deposition rate detected at the QCMs includes a contribution from the gas-phase element, which cannot be distinguished in-situ without additional methods. 

Even for well-behaved cases, limited RBS validation of composition maps will always be advisable when using new target combinations, although it is only necessary to perform on a small subset of samples at the start of an SDL campaign, and thereafter for occasional spot-checks. When producing many samples, it will also be necessary to occasionally retrain GPs for the sputtering targets due to the steady erosion of the target surface, which will change the flux distribution. The need for this can be determined by monitoring target voltages and cross-checking the predictions of the stored GPs with incoming new data, again performed automatically in the SDL in a short time.

% one paragraph discussing ML aspect
There are further possibilities to improve the ML models developed and, in doing so, aid decision making in the SDL. We investigated single-objective GPs with different variance reducing and information-theoretic acquisition functions. To further improve the efficiency and accuracy of the predictions, this could be extended to multi-objective GPs, and the development of customised acquisition functions and kernels tailored for the problem at hand. However we note that in this application, the corresponding time saving would be minor. Other parts of the SDL-workflow could also be fine-tuned to reach optimum conditions. Instead of having a fixed budget of 30 experiments, a more dynamic criterion, based on the mean variance of the GP in relation to the QCM signals, and their own errors, could be implemented.

To summarise, we have developed an ML-driven magnetron sputtering Self-driving lab (SDL) able to accurately synthesise combinatorial thin film samples of some targeted composition and predict quantitive chemical composition maps. We proposed an active learning framework to learn sputter deposition flux across the power and pressure process space for each magnetron source, using in-situ data from several QCM sensors, and we determined initialisation and query strategies to achieve this result using as few queries as possible. We have shown empirically that a fully Bayesian Gaussian process with the BALM acquisition function achieves the best result, being able to learn the deposition rate vs power and pressure in less than 10 experiments. The less computationally-intensive standard GP with a NIPV acquisition function also performs well, and can learn in about 30 queries. Both cases correspond to 15-30 minutes of experimental time to learn a given source. In addition, we have empirically tested and verified that a combination of single-source learning and a geometric sputter-flux model can be used to precisely predict co-sputtering composition maps on produced thin films, as verified by Rutherford backscattering spectroscopy. Besides allowing automating recipe generation for a specified composition, our approach avoids the need to measure a composition map for every sample produced in the SDL. This creates a huge time-saving given that the SDL is capable of producing a new sample every 20 minutes. Our results illustrate the transformative potential of machine learning coupled with physical process models and in-situ characterisation to accelerate materials exploration in self-driving systems. Future extension of the SDL will focus on the conversion of the sputtered multi-element composition-graded thin films into novel functional materials via reactive thermal processing, and ML-guided derivation of their synthesis-property relationships.

\section*{Acknowledgements}
The authors thank Carl Hvarfner for helpful discussions and Corrado Comparotto and Younes Lablali for assisting in RBS measurements. This work is supported by Swedish Foundation for Strategic Research (SSF), the strategic research area STandUP for Energy, the Wallenberg AI, Autonomous Systems and Software Program (WASP) funded by the Knut and Alice Wallenberg Foundation. The work made use of the Myfab clean-room at Uppsala University, part of a VR and KAW funded national infrastructure, and the National Academic Infrastructure for Super computing in Sweden (NAISS). Operation of the accelerator, used for RBS measurements, is supported by the Swedish Research Council VR-RFI (Contracts 2019\_00191 \& 2023\_00155). 

% \bibliography{references}

\newpage
\appendix

\section{Experimental settings}
The following experimental setting were used for the flux model
\begin{itemize}
    \item Angle to magnetron $\varphi\colon 28.4 \degree$ 
    \item Angle to substrate $\theta\colon 43.2 \degree$ 
    \item Distance to substrate $r\colon 155$mm 
\end{itemize}

\begin{figure}
    \centering
    \includegraphics[width=\linewidth]{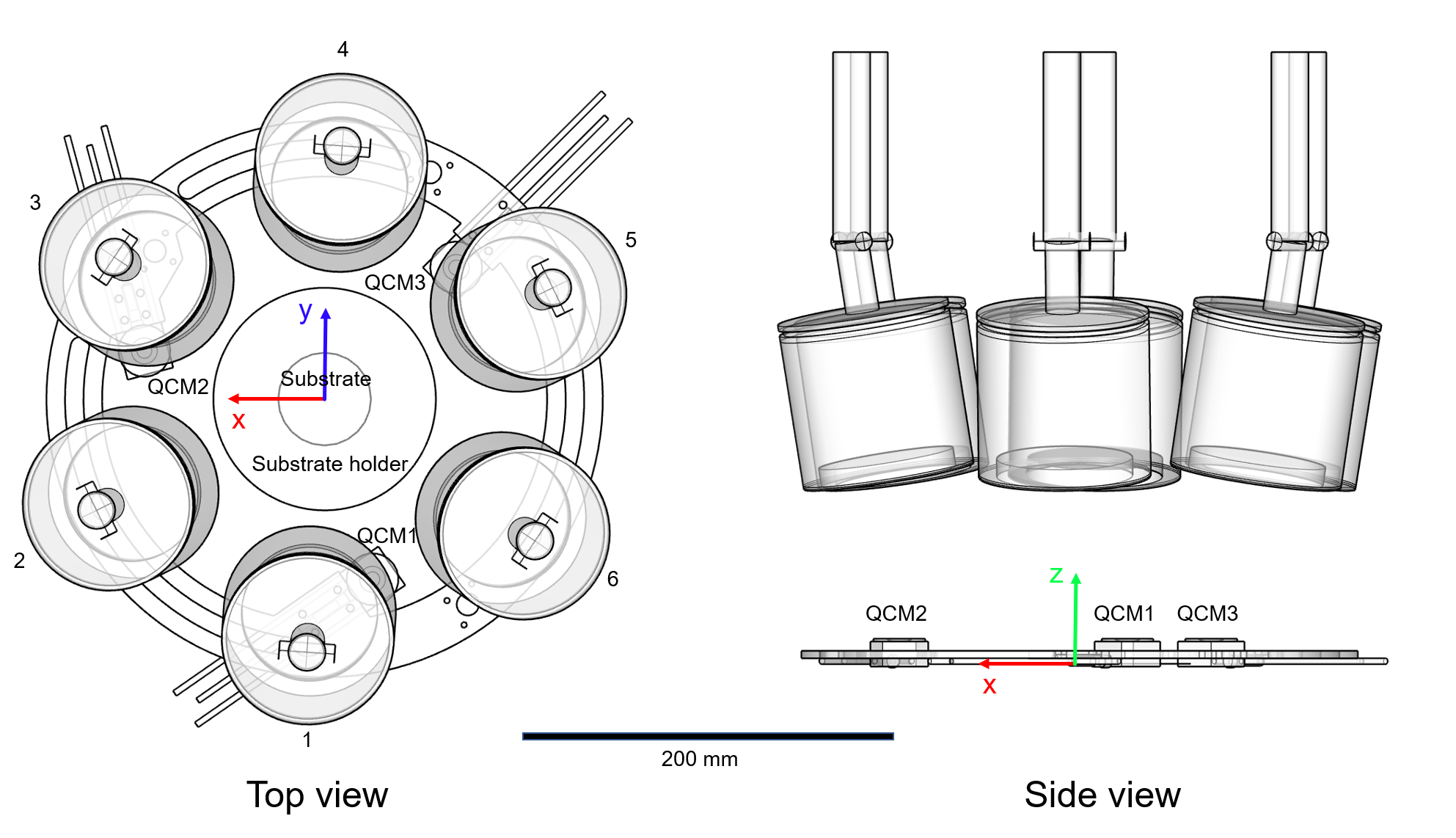}
    \caption{Sputter chamber showing the magnetron and sample placements.}
    \label{fig:supp_sputter_chamber}
\end{figure}

Figure \ref{fig:supp_sputter_chamber} shows a scale diagram of the sputter chamber, where we see how the substrate, QCMs and magnetrons are placed, both from above and the side. 

\section{Additional multi-element verification}

\begin{figure}
    \centering
    \includegraphics[width=\linewidth]{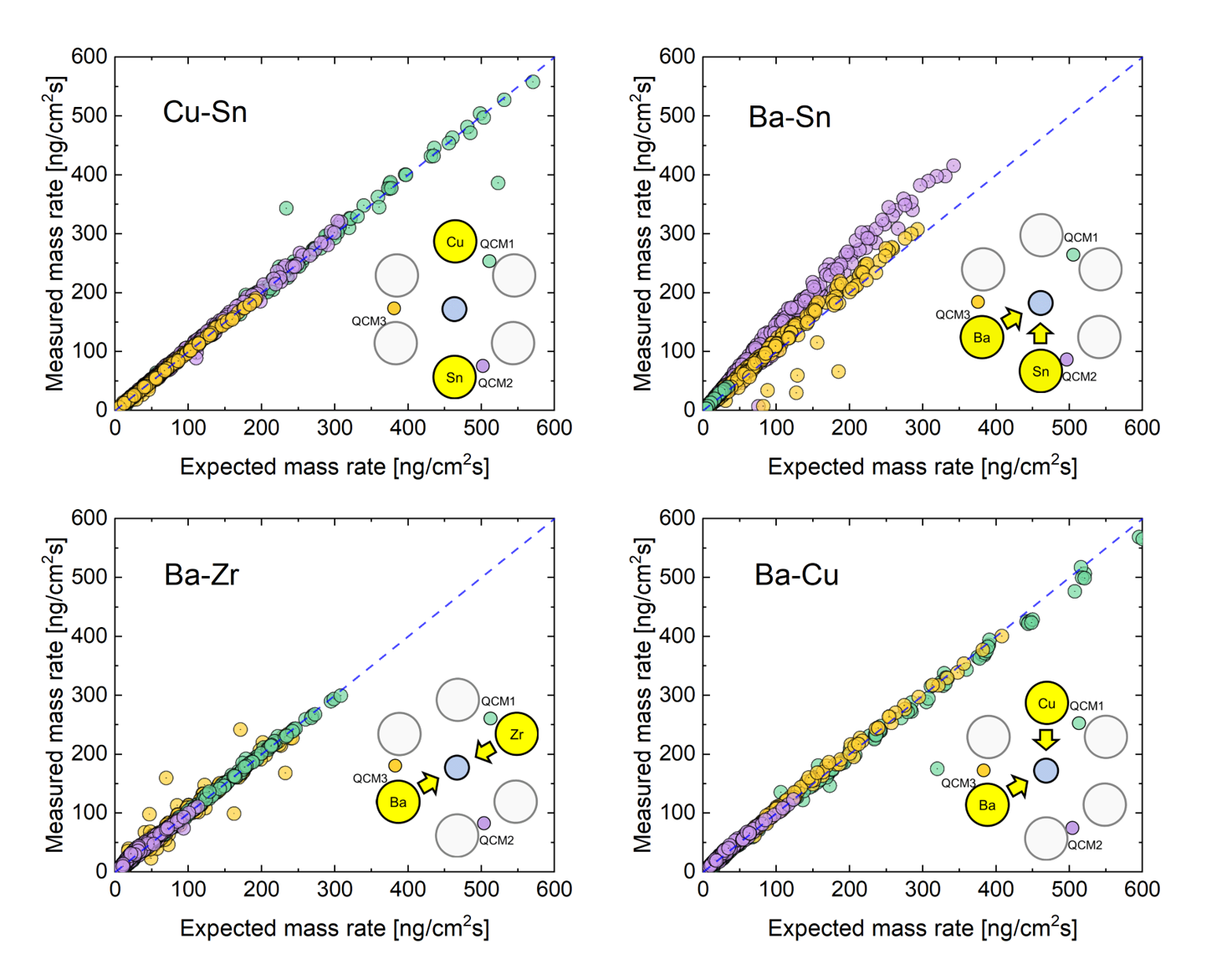}
    \caption{Comparisons of the expected deposition rates based on the sum of two trained GPs (one for each target) and the measured deposition rates obtained when the two targets are sputtered together. The dashed diagonal lines are where points should fall for an accurate prediction. The relevant targets in each figure, their positions and the position of the QCMs are shown in the inset diagrams. Deviations from the expected additive behaviour are seen when the two targets are neighbours, due to plasma-plasma interactions - see for the case of Sn-Ba.}
    \label{fig:supp_2sources}
\end{figure}

\begin{figure}
    \centering
    \includegraphics[width=0.8\linewidth]{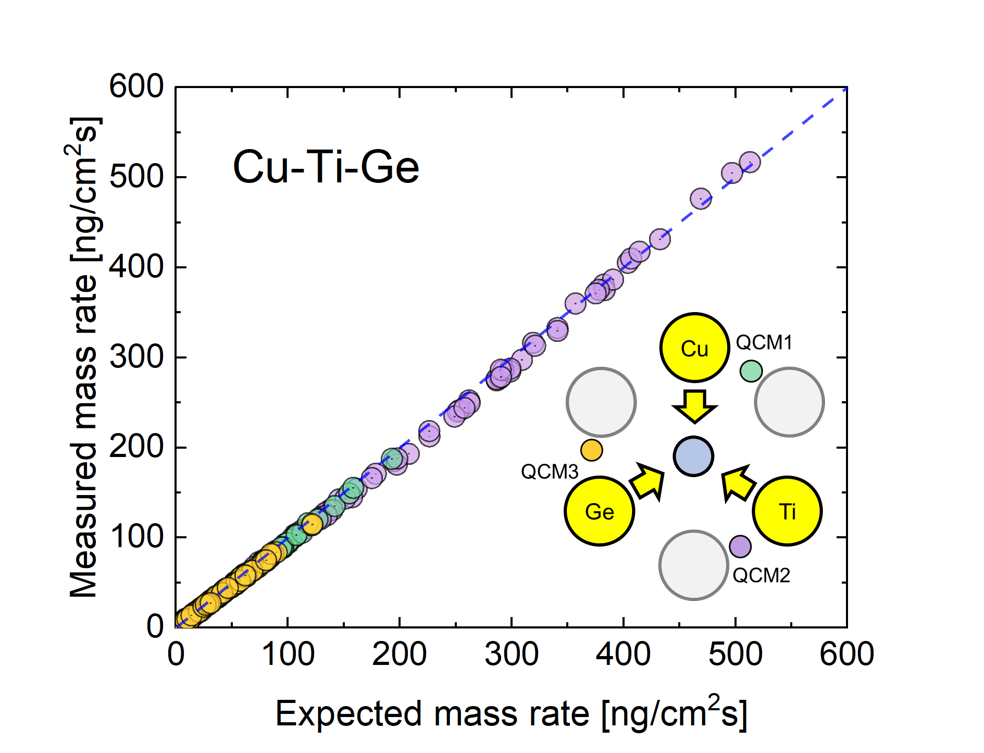}
    \caption{Comparisons of the expected and actual deposition rates for 3 sources, Cu, Ge and Ti. Sputtered with randomized conditions over 100 runs, similar to the analysis in Figure \ref{fig:supp_2sources}}.
    \label{fig:supp_3sources}
\end{figure}

Figure \ref{fig:supp_2sources} and \ref{fig:supp_3sources} compares the expected deposition rate from taking the sum of the GPs from two respectively three different elements, with the actual deposition rate during co-sputtering. The positions of the targets in the sputtering chamber are also indicated in the figure lower right corner.

\begin{figure}
    \centering
    \includegraphics[width=\linewidth]{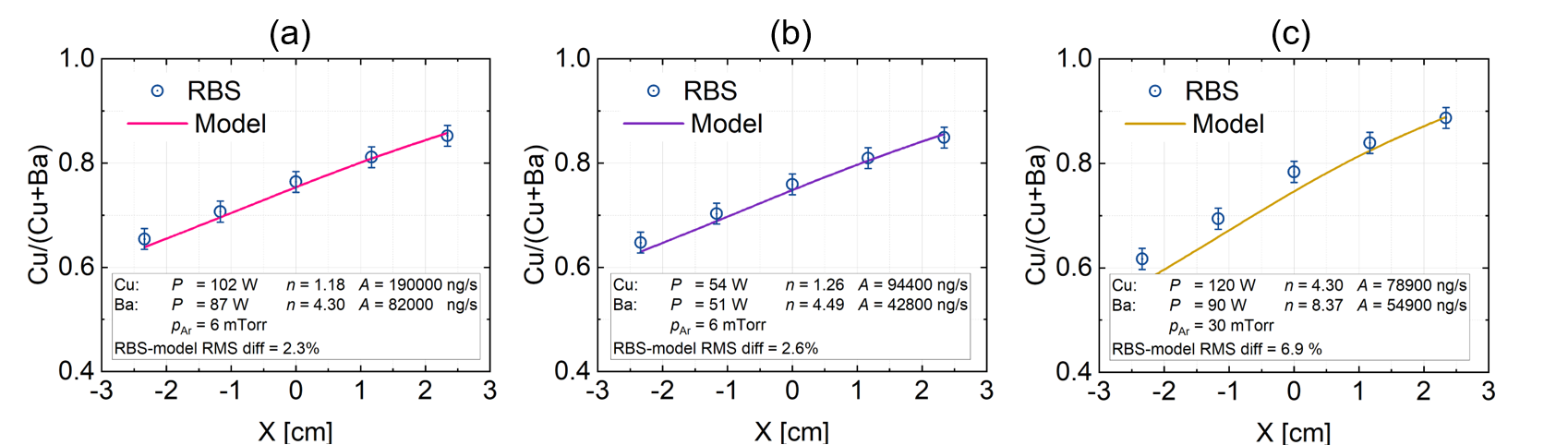}
    \caption{Comparison of RBS measurements and predicted compositions across 3 Cu-Ba samples. Sputtering conditions and model parameters are indicated in the figures.}
    \label{fig:more_rbs}
\end{figure}

Figure \ref{fig:more_rbs} shows the comparison of RBS measurements and predicted compositions across 3 Ba-Cu samples. The sputter conditions, model parameters and prediction error are displayed in each figure.

% \end{document}
\end{document}